\pdfoutput=1
\documentclass[11pt]{article}
\usepackage[final]{acl}
\usepackage{times}
\usepackage{latexsym}
\usepackage{longtable}
\usepackage{tabularray}
\usepackage{colortbl}
\definecolor{1}{RGB}{198, 217, 239}
\definecolor{2}{RGB}{241, 220, 219}
\usepackage[most]{tcolorbox}
\usepackage{float}
\DefTblrTemplate{firsthead,middlehead,lasthead}{default}{
}
\DefTblrTemplate{firstfoot}{default}{
  \UseTblrTemplate{contfoot}{default}
  \UseTblrTemplate{caption}{default}
}
\DefTblrTemplate{middlefoot}{default}{
  \UseTblrTemplate{contfoot}{default}
  \UseTblrTemplate{capcont}{default}
}
\ExplSyntaxOn
\DefTblrTemplate{lastfoot}{default}{
  \UseTblrTemplate{note}{default}
  \UseTblrTemplate{remark}{default}
  \int_compare:nNnTF { \l__tblr_table_page_int } = {1} {
    \UseTblrTemplate{caption}{default}
  } {
    \UseTblrTemplate{capcont}{default}
  }
}
\ExplSyntaxOff
\usepackage[T1]{fontenc}
\usepackage[utf8]{inputenc}
\usepackage{microtype}
\usepackage{inconsolata}
\usepackage{nicematrix}
\usepackage{graphicx}
\usepackage{multirow}
\usepackage{multicol}
\usepackage{cleveref}
\usepackage{paralist}
\usepackage{booktabs}
\usepackage{rotating}
\usepackage{enumitem}
\usepackage{tabularray}
\usepackage{array}
\usepackage{xcolor}
\usepackage[misc]{ifsym}
\title{
    Are LLMs Good Zero-Shot Fallacy Classifiers?
}

\author{
  Fengjun Pan$^{1,2}$\thanks{Equal contribution.~~\Letter~~Corresponding Author.} \qquad Xiaobao Wu$^2$\footnotemark[1] \qquad Zongrui Li$^{1}$ \qquad Anh Tuan Luu$^{2}$\textsuperscript{\Letter} \\
  $^1$Interdisciplinary Graduate Programme, Nanyang Technological University, Singapore\\
  $^2$College of Computing and Data Science, Nanyang Technological University, Singapore\\
  \texttt{panf0004@e.ntu.edu.sg} \qquad
  \texttt{xiaobao002@e.ntu.edu.sg} \\
  \texttt{zongrui001@e.ntu.edu.sg} \qquad
  \texttt{anhtuan.luu@ntu.edu.sg}
}

\def\eg{\emph{e.g.}} 
\def\ie{\emph{i.e.}} 

\def\etc{\emph{etc.}}
\usepackage{makecell}
\begin{document}
\maketitle

\begin{abstract}
Fallacies are defective arguments with faulty reasoning.
Detecting and classifying them is a crucial NLP task to prevent misinformation, manipulative claims, and biased decisions.
However, existing fallacy classifiers are limited by the requirement for sufficient labeled data for training,
which hinders their out-of-distribution ({OOD}) generalization abilities.
In this paper, we focus on leveraging Large Language Models (LLMs) for zero-shot fallacy classification.
To elicit fallacy-related knowledge and reasoning abilities of LLMs,
we propose diverse single-round and multi-round prompting schemes, applying different task-specific instructions such as extraction, summarization, and Chain-of-Thought reasoning.
With comprehensive experiments on benchmark datasets,
we suggest that LLMs could be potential zero-shot fallacy classifiers. In general, LLMs under single-round prompting schemes have achieved acceptable zero-shot performances compared to the best full-shot baselines and can outperform them in all OOD inference scenarios and some open-domain tasks. Our novel multi-round prompting schemes can effectively bring about more improvements, especially for small LLMs. 
Our analysis further underlines the future research on zero-shot fallacy classification.
Codes and data are available at:
{\hypersetup{urlcolor=magenta}\url{https://github.com/panFJCharlotte98/Fallacy_Detection}}.
\end{abstract}

\section{Introduction}
\begin{figure}
  \vspace{-10pt}
  \centering
  \includegraphics[width=\linewidth]{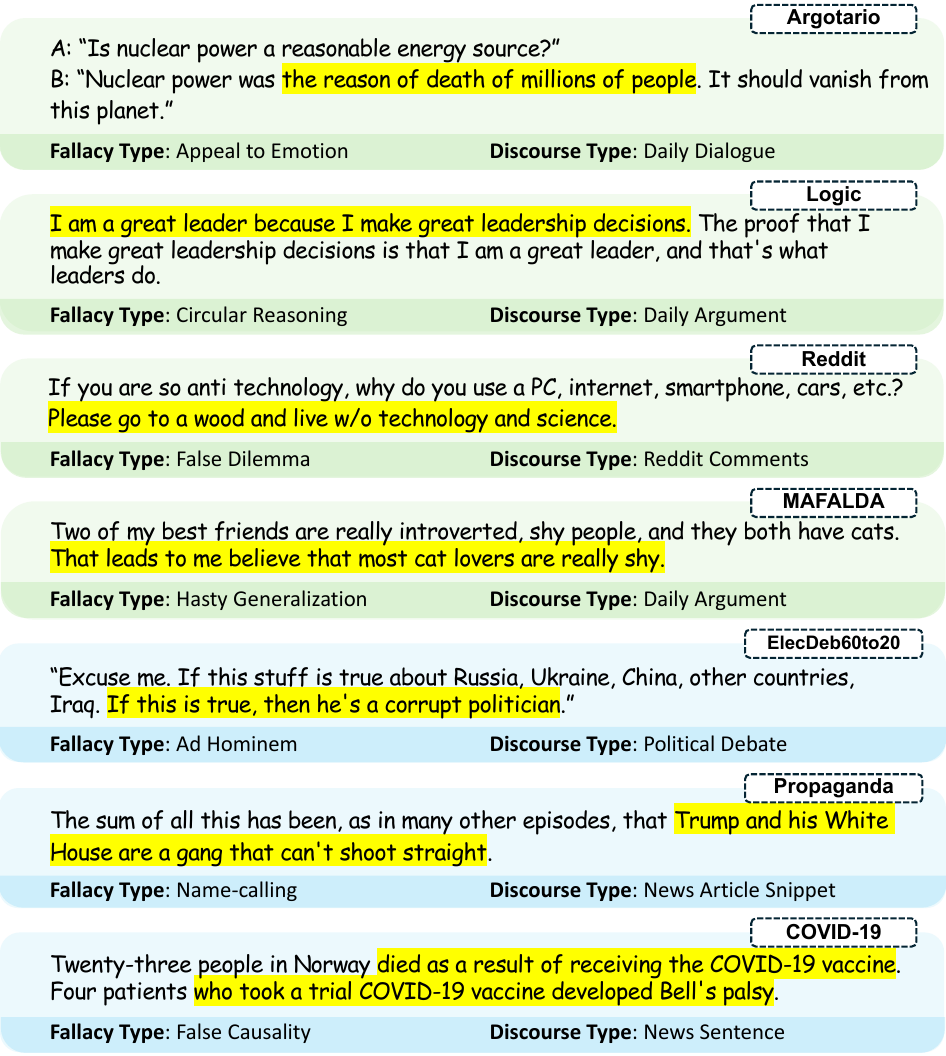}
  \caption{
    Examples of fallacies and their types from existing datasets.
  }
  \label{fig_examples}
  \vspace{-20pt}
\end{figure}

    A fallacy is a defective argument derived from erroneous or invalid reasoning that may appear to be reasonable but are, in fact, logically unsound or faulty \cite{woods2004cares,damer2008attacking,van2009fallacies,hamblin2022fallacies}.
    \Cref{fig_examples} illustrates examples of different fallacy types,
    for instance, a fallacy type of Circular Reasoning: \emph{I am a great leader because I make great leadership decisions.}
    Fallacies commonly appear in various scenarios, such as news articles \cite{news-fallacy}, advertisements \cite{advertising}, propaganda \cite{walton1997propaganda}, politics \cite{politics},
    and social media \cite{social-media}.
    They could be intentionally exploited to disseminate misinformation \cite{misinfo-fallacy}, manipulate public opinions, undermine rational discussions, and influence critical decision-making \cite{visserreason}.
    In consequence, detecting and classifying fallacies becomes an imperative challenge.

    However, existing fallacy classification methods typically follow a full-shot supervised fine-tuning manner,
    limited by three issues:
    \begin{inparaenum}[(\bgroup\bfseries i\egroup)]
        \item
            They require sufficient labeled data for training \cite{dataset-elecdebate,t5-multitask-fallacy-detection},
            but collecting these data is often time-consuming and expensive.
            This is because annotating fallacy data typically rely on expert knowledge,
            due to the complex and esoteric nature of fallacies \cite{benchmark-mafalda}.
        \item
            They cannot well generalize to out-of-distribution (OOD) fallacies and discourse types \cite{ood_problem}
            due to the inherent limitation of supervised learning.
            Once given an unseen fallacy class or discourse from other domains, they have to retrain a new model from scratch, which costs considerable computational and time resources.
        \item
            They are susceptible to imbalanced data \cite{alhindi-gen-fallacy-examples}.
            They could reach high performance on the dominant fallacy types while low on infrequent ones.
    \end{inparaenum}
    As a result, full-shot fallacy classifiers cannot fulfill real-world application scenarios.

    Motivated by the above issues, in this paper, we explore zero-shot fallacy classification, \ie, classifies fallacies without training data.
    Specifically, we concentrate on Large Language Models (LLMs) as they have been extensively pretrained and possess wide knowledge and strong reasoning abilities.
    To investigate LLMs' performance on this task,
    we consider two kinds of prompting schemes.
    First, we employ a \textbf{single-round prompting scheme}.
    We simply prompt LLMs to classify fallacies either with or without manually crafted fallacy type definitions.

    Second, to elicit the inherent fallacy-related knowledge and reasoning abilities of LLMs, we further propose diverse \textbf{multi-round prompting schemes}:
    we instruct LLMs to analyze and classify fallacies through definition generation, general fallacy analysis (with warm-up), premise \& conclusion analysis, and Chain-of-Thought.
    Under these prompting schemes,
    we observe that zero-shot prompted LLMs can outperform or achieve comparable performances with SOTA full-shot fine-tuned T5 baselines on some open-domain benchmark datasets and can generally achieve sub-optimal performances on hard domain-specific datasets.
    More importantly, LLMs consistently demonstrate advantage over full-shot trained baselines on OOD inferences on low-resource fallacy classification tasks.
    Besides, our novel multi-round prompting schemes can effectively improve LLMs' classification performance compared to the single-round prompts, especially for small LLMs, \eg, Llama3, Qwen2.5 and Mistral.
    We conclude our main contributions as below:
    \begin{itemize}[leftmargin=*,itemsep=0pt]
        \item
            We propose diverse novel prompting schemes, including both the basic single-round and the advanced multi-round ones, 
            that are sufficiently effective in eliciting the fallacy-related knowledge and reasoning abilities of LLMs for zero-shot fallacy classification.
        \item
            We are the first to conduct extensive experiments with representative LLMs on existing fallacy benchmark datasets and provide overviews and insights concerning the boundary of LLMs' zero-shot fallacy classification performances. 
            We show that we can resort to LLMs as potential zero-shot fallacy classifiers that solely rely on LLMs' inherent knowledge without computationally intensive training full-shot models from scratch.
        \item
            We provide detailed analysis on LLMs, datasets, and prompting schemes and summarize a general guidance for choosing the potentially most effective prompting schemes for different fallacy classification scenarios with respect to data domains and LLM types, inspiring future research on zero-shot fallacy classification and other relevant linguistic reasoning tasks.
    \end{itemize}

\begin{figure*}[t]
    \vspace{-10pt}
      \centering
      \includegraphics[width=\linewidth]{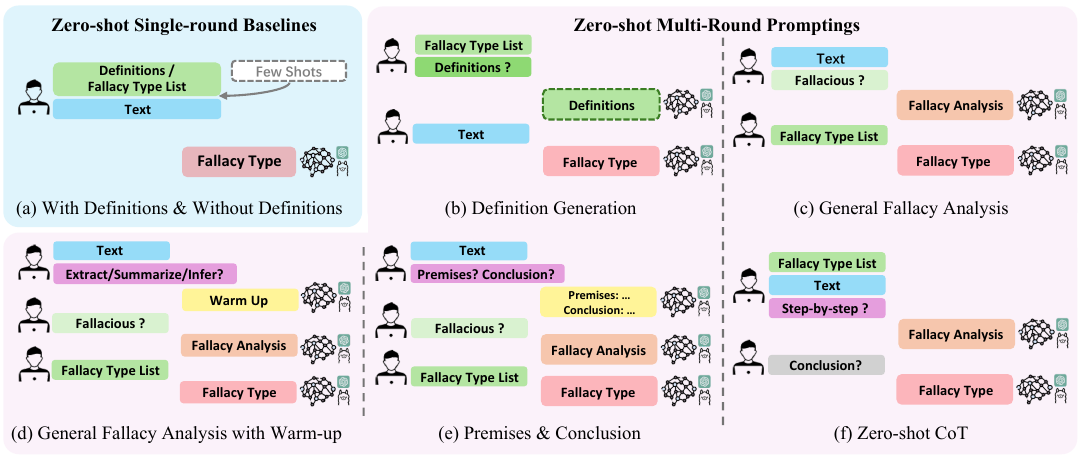}
      \vspace{-20pt}
      \caption{
        Illustration of single-round and multi-round prompting schemes.
        (\textbf{a}):
            Prompt LLMs to classify with or without fallacy type definitions. 
        (\textbf{b}):
            Prompt LLMs to generate fallacy type definitions and then classify.
        (\textbf{c}):
            Prompt LLMs to analyze the input discourse and then classify.
        (\textbf{d}):
            Prompt LLMs to warm up (extract, summarize, or infer), analyze the input discourse, and then classify.
        (\textbf{e}):
            Prompt LLMs to extract the premises and conclusion, analyze the input discourse, and then classify.
        (\textbf{f}):
            Prompt LLMs to reason step by step to classify and then draw an answer.
      }
      \label{fig_method}
\end{figure*}

\section{Related Work}
\paragraph{Fallacy Classification}
    Fallacy detection and classification is an emerging natural language processing task that has received increasing attention.
    Many fallacy datasets across different discourse genres and various domains have been created in a line of work~\cite{dataset-argotario, before_name_calling, dataset-propaganda, dataset-reddit, dataset-elecdebate, dataset-logic, t5-multitask-fallacy-detection, dataset-covid, benchmark-mafalda}. 
    Early fallacy classification methods are mainly based on traditional machine learning models \cite{Wu2020short,wu2022mitigating,wu2023infoctm,wu2024traco,wu2024topmost,wu2024fastopic,wu2024dynamic,wu2024survey}, \eg, SVM, Bi-LSTM~\cite{habernal2018adapting} and CNN~\cite{before_name_calling}.
    Deep learning methods based on models with Transformers architecture \eg, BERT~\cite{devlin2018bert}, RoBERTa~\cite{liu2019roberta}, T5~\cite{t5} have been proposed in recent works~\cite{dataset-propaganda,dataset-reddit,elecdebate,dataset-logic,t5-multitask-fallacy-detection}. 
    However, all these methods are trained in a supervised manner that is highly reliant on the availability and sufficiency of labeled fallacy data, rendering them struggling on OOD fallacy and discourse types.
    Recently, LLMs' abilities related to fallacy understanding and detection have been explored~\cite{llms-susceptible-fallacies,alhindi-gen-fallacy-examples,sutd-llms-fallacy,logic-translate-for-fallacy-detection, most-relevant} but not evaluated in-depth considering prompt techniques. 
    
\paragraph{Large Language Models}
    The ability of LLMs in logical reasoning has been quantitatively or qualitatively evaluated,
    \eg, commonsense causality reasoning~\cite{kiciman2023causal,talmor2020leap,willig2022can} and abstract reasoning~\cite{gendron2023large,pan2023fact,wu2024updating}.
    Particularly, evidence~\cite{bubeck2023sparks} has verified the existence of inconsistency in its reasoning process (same for the latest LLMs like GPT-4).
    This motivates us to guide LLMs through the proposed multi-round prompting schemes.
    Our work distinguishes from the previous in two points:
    (\textbf{i}) instead of a general reasoning task, we focus on the fallacy classification, a more advanced reasoning task that requires not only expert knowledge of 
    but also rigorous logical reasoning.
    (\textbf{ii}) Apart from the common single-round prompting,
    we propose the novel multi-round prompting schemes that bring along noticeable improvements for small LLMs.

\section{Methodology}
    In this section, we apply LLMs for zero-shot fallacy classification
    by two kinds of prompting schemes: \textbf{Zero-shot Single-Round Prompting} and \textbf{Zero-shot Multi-Round Prompting}.

\subsection{Zero-shot Single-Round Prompting} \label{sec:4.1}

    We first propose zero-shot single-round prompting schemes.
    Following the state-of-the-art fine-tuning baseline \cite{t5-multitask-fallacy-detection},
    the single-round prompt consists of three components, as shown in \Cref{fig_method} (a):
    \begin{inparaenum}[(i)]
        \item
            A label space that defines a limited number of fallacy types;
        \item
            An input fallacious discourse to be classified;
        \item
            The task and output format instructions.
    \end{inparaenum}
    For the label space, we consider two settings:
    \begin{inparaenum}[(i)]
        \item
            \textbf{Without Definitions}, which merely enumerates all the viable fallacy types as an option list,
            for example, \textit{1. Appeal to Emotion, 2. Ad Hominem, 3. False Dilemma ...}.
        \item
            \textbf{With Definitions}, which means that we provide the natural language definition of each fallacy type,
            for example, \textit{1. Appeal to Emotion is a fallacy when someone attempts to argue or persuade by using emotive language to arouse non-rational sentiments within the intended audience. 2. Ad Hominem is a fallacy when someone attacks the others' characters or motives instead of addressing the substance of their arguments...}
    \end{inparaenum}
    See the used single-round prompt templates in \Cref{tab:prompt_template}.

\subsection{Zero-shot Multi-Round Prompting}
    Besides, we propose zero-shot multi-round prompting schemes.
    This is motivated by the complexity nature of the fallacy classification task
    as it involves reading comprehension, information extraction, logical reasoning, knowledge recall, and pattern recognition.
    Therefore, we introduce the following multi-round prompting schemes that aim to elicit LLMs' inherent knowledge and reasoning abilities on fallacy.
    \Cref{fig_method} summarizes all these schemes.

\paragraph{Definition Generation}
    As illustrated in \Cref{fig_method} (b),
    we prompt LLMs to generate the definition for each fallacy given in the fallacy type list in the first round,
    and then classify the fallacy type of the input discourse based on these definitions in the second round.
    This is because LLMs probably have learned the knowledge about different fallacy types during pretraining.
    This scheme elicits LLMs to recall these knowledge.
    Besides, the generated definitions better align with LLMs' understanding compared to manually crafted definitions
    due to their auto-regressive paradigm, thus may enhance the reasoning process of determining fallacy types.

\paragraph{General Fallacy Analysis}
    As shown in \Cref{fig_method} (c),
    we first instruct LLMs to analyze the input discourse and determine whether it is logically reasonable or potentially fallacious;
    then we ask LLMs to determine the fallacy type.
    The first round works as an intermediate step to offer analytical information for any detected potential fallacy, which assists the fallacy classification in the second round.
    Note that here we use a neutral instruction which includes both positive (logically reasonable) and negative (potentially fallacious) possibilities to avoid any biased implication.

\paragraph{General Fallacy Analysis with Warm Up}
    In this scheme (\Cref{fig_method} (d)), we add a warm-up round as the first round that asks LLMs to extract, paraphrase, summarize or infer about the content and context of the input discourse, and then follow the above General Fallacy Analysis scheme.
    This scheme is inspired by the fact that discourses in some domain-specific datasets are truncated, which greatly hinders LLMs' understanding.
    The warm-up round brings more contextual information to ease LLMs' understanding, which thus benefits their predictions on fallacy types.

\paragraph{Premises \& Conclusion}
    \citet{benchmark-mafalda} give a formal definition of the term ``fallacy'': \textit{{A fallacy is an argument where the premises do not entail the conclusion}}.
    Following this formal definition, we use three rounds as shown in \Cref{fig_method} (e):
    First, we prompt LLMs to extract the premises and conclusion of the input discourse.
    Second,
    we ask LLMs to determine whether the input is fallacious by analyzing if the premises entail the conclusion according to the formal definition.
    Finally, LLMs predict the fallacy type.

\paragraph{Zero-shot CoT}
    Lastly, we consider zero-shot CoT (Chain-of-Thought).
    Previous multi-round prompting schemes all apply task-specific instructions in intermediate steps.
    Differently as shown in \Cref{fig_method} (f) zero-shot CoT directly leverages the magic instruction {\it {Now, let’s think step by step}} \cite{kojima2022large}
    to prompt LLMs to classify fallacy types through step-by-step reasoning.
    This elicits LLMs to derive analogous reasoning chains \cite{cotsurvey}.

\begin{table}[!t]
  \centering
  \setlength{\tabcolsep}{1mm}
  \renewcommand{\arraystretch}{1.1}
  \resizebox{\linewidth}{!}{
    \begin{tabular}{llrrl}
    \toprule
    \textbf{Domain} & \textbf{Dataset} & \makecell[c]{\textbf{Splits} \\ (train/dev/test)} & \textbf{\#FT.} & \textbf{DT.} \\
    \midrule
    \multirow{4}[1]{*}{Open} & {\sc Argotario} & 909/102/312 & 5     & QA \\
          & {\sc Logic} & 1849/300/300 & 13    & Unlimited \\
          & {\sc Reddit} & 1195/342/513 & 8     & Reddit Comments \\
          & {\sc Mafalda} & 0/0/200 & 23    & Unlimited \\
    \midrule
    Politics & {\sc ElecDeb} & 1267/136/154 & 5     & Debate Transcripts \\
    News  & {\sc Propaganda} & 1583/265/265 & 13    & News Articles \\
    COVID News & {\sc COVID-19} & 0/0/154 & 9     & News \& Posts \\
    \bottomrule
    \end{tabular}%
    }
    \caption{Fallacy dataset statistics. \textbf{\#FT.}: Number of fallacy types. \textbf{DT.}: Discourse type.}
  \label{tab:dataset_statistics}%
\end{table}%
\begin{table*}[!ht]
  \centering
  \resizebox{\linewidth}{!}{
    \begin{tabular}{cclrrrrrrr}
    \toprule
    \textbf{Shot} & \textbf{Round} & \multicolumn{1}{c}{\textbf{Model}} & \multicolumn{1}{c}{\textsc{\textbf{Argotario}}} & \multicolumn{1}{c}{\textsc{\textbf{Logic}}} & \multicolumn{1}{c}{\textsc{\textbf{Reddit}}} & \multicolumn{1}{c}{\textsc{\textbf{ElecDeb}}} & \multicolumn{1}{c}{\textsc{\textbf{Propaganda}}} & \multicolumn{1}{c}{\textsc{\textbf{Mafalda}}} & \multicolumn{1}{c}{\textsc{\textbf{Covid-19}}} \\
    \midrule
    \multirow{6}[4]{*}{Full} & \multirow{6}[4]{*}{N/A} & T5-3B Single-task & 69.13 & \textbf{64.95} & \textbf{83.20} & \textbf{62.37} & 38.36 & -     & - \\
          &       & T5-3B Multi-ALR & \textbf{72.38} & 63.54 & 81.88 & \cellcolor[rgb]{ .773,  .851,  .945}33.22 & \cellcolor[rgb]{ .773,  .851,  .945}12.58 & \cellcolor[rgb]{ .773,  .851,  .945}31.52 & \cellcolor[rgb]{ .773,  .851,  .945}12.28 \\
          &       & T5-3B Multi-ALEP & 70.51 & 61.65 & \cellcolor[rgb]{ .773,  .851,  .945}56.98 & 56.35 & \textbf{43.33} & \cellcolor[rgb]{ .773,  .851,  .945}\textbf{35.60} & \cellcolor[rgb]{ .773,  .851,  .945}\textbf{14.59} \\
\cmidrule{3-10}          &       & T5-large Single-task & 58.26 & 55.23 & 77.77 & 41.48 & 38.62 & -     & - \\
          &       & T5-large Multi-ALR & 65.65 & 59.48 & 80.42 & \cellcolor[rgb]{ .773,  .851,  .945}37.22 & \cellcolor[rgb]{ .773,  .851,  .945}8.82 & \cellcolor[rgb]{ .773,  .851,  .945}25.13 & \cellcolor[rgb]{ .773,  .851,  .945}13.07 \\
          &       & T5-large Multi-ALEP & 64.14 & 57.67 & \cellcolor[rgb]{ .773,  .851,  .945}38.87 & 56.15 & 39.75 & \cellcolor[rgb]{ .773,  .851,  .945}25.60 & \cellcolor[rgb]{ .773,  .851,  .945}14.08 \\
    \midrule
    \multirow{14}[4]{*}{Zero} & \multirow{7}[2]{*}{Single} & GPT-4 & 78.94 & 50.43 & \cellcolor[rgb]{ .949,  .863,  .859}79.10 & \cellcolor[rgb]{ .949,  .863,  .859}42.26$^{\circ}$ & \cellcolor[rgb]{ .949,  .863,  .859}34.8$^{\circ}$ & \cellcolor[rgb]{ .949,  .863,  .859}48.74$^{\circ}$ & \cellcolor[rgb]{ .949,  .863,  .859}20.47 \\
          &       & GPT-3.5 & 63.59$^{\circ}$ & 39.65 & \cellcolor[rgb]{ .949,  .863,  .859}70.42$^{\circ}$ & \cellcolor[rgb]{ .949,  .863,  .859}41.01 & \cellcolor[rgb]{ .949,  .863,  .859}22.39 & 31.27 & \cellcolor[rgb]{ .949,  .863,  .859}17.45 \\
          &       & Qwen2.5-Instruct 14B & 68.19$^{\circ}$ & 41.82$^{\circ}$ & \cellcolor[rgb]{ .949,  .863,  .859}67.72$^{\circ}$ & \cellcolor[rgb]{ .949,  .863,  .859}37.28 & \cellcolor[rgb]{ .949,  .863,  .859}21.89 & 33.03$^{\circ}$ & \cellcolor[rgb]{ .949,  .863,  .859}17.6 \\
          &       & Qwen2.5-Instruct 7B & 61.38$^{\circ}$ & 35.48$^{\circ}$ & \cellcolor[rgb]{ .949,  .863,  .859}58.58 & \cellcolor[rgb]{ .949,  .863,  .859}43.34 & \cellcolor[rgb]{ .949,  .863,  .859}16.03 & 31.27$^{\circ}$ & \cellcolor[rgb]{ .949,  .863,  .859}15.19$^{\circ}$ \\
          &       & Llama3-Chat 8B & 48.87 & 27.45$^{\circ}$ & 49.41 & \cellcolor[rgb]{ .949,  .863,  .859}39.36$^{\circ}$ & \cellcolor[rgb]{ .949,  .863,  .859}17.30 & 24.85$^{\circ}$ & 11.00$^{\circ}$ \\
          &       & Mistral-Instruct 7B & 57.04$^{\circ}$ & 28.99$^{\circ}$ & 46.89$^{\circ}$ & 33.23$^{\circ}$ & \cellcolor[rgb]{ .949,  .863,  .859}16.89$^{\circ}$ & 23.44$^{\circ}$ & \cellcolor[rgb]{ .949,  .863,  .859}14.69 \\
          &       & Llama2-Chat 13B & 50.20  & 25.11 & 34.15$^{\circ}$ & 35.57 & 10.61 & 22.09 & 14.15 \\
\cmidrule{2-10}          & \multirow{7}[2]{*}{Multi} & GPT-4 & \textcolor[rgb]{ 1,  0,  0}{\textbf{79.87}} & \textcolor[rgb]{ 1,  0,  0}{\textbf{50.54}} & \textcolor[rgb]{ 1,  0,  0}{\textbf{81.11}} & 41.25 & \textcolor[rgb]{ 1,  0,  0}{\textbf{35.37}} & \textcolor[rgb]{ 1,  0,  0}{\textbf{52.86}} & \textcolor[rgb]{ 1,  0,  0}{\textbf{25.18}} \\
          &       & GPT-3.5 & 68.40 & 41.11 & 71.08 & 37.77 & 26.67 & 40.73 & 17.24 \\
          &       & Qwen2.5-Instruct 14B & 68.87 & 45.89 & 77.08 & 34.09 & 26.60  & 45.94 & 23.73 \\
          &       & Qwen2.5-Instruct 7B & 60.20  & 40.09 & 64.58 & \textcolor[rgb]{ 1,  0,  0}{\textbf{44.55}} & 19.22 & 35.37 & 22.88 \\
          &       & Llama3-Chat 8B & 61.39 & 35.66 & 57.83 & 40.81 & 21.35 & 34.18 & 19.83 \\
          &       & Mistral-Instruct 7B & 57.26 & 31.43 & 59.70  & 32.91 & 20.41 & 29.08 & 18.53 \\
          &       & Llama2-Chat 13B & 48.79 & 28.85 & 45.82 & 36.37 & 11.11 & 15.68 & 14.16 \\
    \bottomrule
    \end{tabular}%
    }
    \caption{
        Fallacy classification results of Macro-F1.
        The best results obtained by T5 baselines are in \textbf{bold}, and our best zero-shot results obtained by LLMs are in \textcolor[rgb]{ 1,  0,  0}{\textbf{bold}}.
        \colorbox{1}{Blue} denotes out-of-distribution (OOD) results of T5 baselines while
        \colorbox{2}{Red} denotes the corresponding zero-shot results where single-round prompted LLMs outperform the OOD results of T5 baselines;
        $\circ$ denotes single-round prompting without definitions, while other single-round results are with definitions.
        Here we only report the best zero-shot results of LLMs across all prompting schemes.
        See \Cref{sec:experiment_result_details} for detailed results.
    }
  \label{tab_main_results}%
\end{table*}%
\section{Experiments}
In this section,
we conduct extensive experiments and evaluate model performance by \textbf{Macro F1} following \citet{t5-multitask-fallacy-detection}.

\paragraph{Datasets.} We consider the following 7 benchmark datasets:
    \begin{inparaenum}[(\bgroup\bfseries i\egroup)]
        \item
            {\sc\textbf{Argotario}} \citep{dataset-argotario} contains short QA pairs about various open topics. The answers may commit one of the 5 fallacy types or no fallacy.
        \item
            {\sc \textbf{Logic}} \citep{dataset-logic} consists of fallacy examples in shot statements or conversations collected from education websites across 13 different fallacy types.
        \item
            {\sc \textbf{Reddit}} \citep{dataset-reddit} collects Reddit comments submitted for different discussion topics. Each data example contains one of the 8 types of fallacies that may occur in multiple spans. Since we find this dataset to be the most balanced one, we combine its validation and test set as the inference set and hold it as an OOD dataset when testing the full-shot baseline.
        \item
            {\sc \textbf{Mafalda}} \citep{benchmark-mafalda} merges and unifies previous fallacy datasets and annotate 200 examples in sentence level with a total of 23 fallacy types. As this dataset contains data examples with multi-labels across different sentences, we take the most dominant fallacy type with most occurrences within each example as the single fallacy label to adhere the unified multi-class single-label classification setup. We also consider this dataset to be OOD as it has insufficient instances for fine tuning.
        \item
            \textsc{\textbf{ElecDeb60to20}} (\textsc{\textbf{ElecDeb}} hereinafter, \citealp{dataset-elecdebate}) proposes a fallacy corpus of political speeches in U.S. presidential election debates. Each data instance is a shot snippet that contains one type of fallacies. We keep 5 fallacy types and remove the examples of {\it Slogans} from the original dataset as it may not strictly align with the definition of a fallacy as discussed in \cite{benchmark-mafalda}. 
        \item
            {\sc \textbf{Propaganda}} \citep{dataset-propaganda} is a large corpus of propaganda techniques used in news articles annotated at sentence level. We exclude 5 propaganda-oriented classes ({\it Loaded Language}, {\it Exaggeration or Minimisation}, {\it Thought-terminating Cliches}, {\it Slogans}, {\it Repetition}) and keep a total of 13 fallacy types. We include the four most adjacent sentences before and after each annotated fallacious sentence as its surrounding context.
        \item
            {\sc \textbf{Covid-19}} \citep{dataset-covid} is a fallacy corpus of news sentences and media posts about COVID pandemic. Each data examples could be non-fallacious or contain one of the 9 types of fallacies. Considering its small size, we combine all the data instances in three available splits as the inference set and hold it as OOD for full-shot baseline.
    \end{inparaenum}
    \Cref{tab:dataset_statistics} summarizes the dataset statistics.
    Fallacy class distribution in each dataset can be found in \Cref{tab:dataset_details}.

\begin{table}[!t]
  \centering
  \renewcommand{\arraystretch}{1.1}
  \resizebox{\linewidth}{!}{
    \begin{tabular}{lrrr}
    \toprule
    \textbf{Scheme} & \textbf{\#R} & \textbf{Rank} & \textbf{\%Failed} \\
    \midrule
    General Fallacy Analysis with Warm Up & 3     & 4.13  & 12.27 \\
    Zero-shot CoT & 2     & 4.33  & 5.99 \\
    General Fallacy Analysis & 2     & 4.35  & 15.09 \\
    With Definitions & 1     & 4.65  & 4.10 \\
    Definition Generation & 2     & 4.77  & 2.85 \\
    Without Definitions & 1     & 4.90  & 5.04 \\
    Premises \& Conclusion & 3     & 5.84  & 12.46 \\
    \bottomrule
    \end{tabular}%
    }
    \caption{Overall rankings on Macro-F1 of multi-round prompting schemes. \textbf{\#R}: Number of rounds.}
  \label{tab_scheme_rank}%
\end{table}%

\begin{table}[t]
  \centering
  \resizebox{\linewidth}{!}{
    \begin{tabular}{cclrrr}
    \toprule
    \textbf{Domain} & \textbf{Base Model} & \textbf{Scheme} & \textbf{\#R} & \textbf{Rank} & \textbf{\%Failed} \\
    \midrule
    \multirow{4}[4]{*}{Open} & \multirow{2}[2]{*}{GPT} & DG & 2     & 2.88  & 3.28 \\
          &       & WD & 1     & 3.25  & 4.00 \\
\cmidrule{2-6}          & \multicolumn{1}{c}{\multirow{2}[2]{*}{\makecell{Small\\($\leq$14B)}}} & GFA-W & 3     & 2.85     & 11.18 \\
          &       & CoT & 2     & 3.79  & 5.19 \\
    \midrule
    \multirow{4}[4]{*}{Specific} & \multirow{2}[2]{*}{GPT} & WD & 1     & 2.50   & 3.65 \\
          &       & CoT & 2     & 3.33  & 7.93 \\
\cmidrule{2-6}          & \multicolumn{1}{c}{\multirow{2}[2]{*}{\makecell{Small\\($\leq$14B)}}} & CoT & 2     & 3.20  & 7.57 \\
          &       & GFA-W & 3     & 3.80  & 15.91 \\
    \bottomrule
    \end{tabular}%
    }
    \caption{Best two zero-shot prompting schemes for different base models and data domains based on average Macro-F1 rankings. DG: Definition Generation. CoT: Zero-shot CoT. GFA-W: General Fallacy Analysis with Warm Up. GFA: General Fallacy Analysis. WD: With Definitions. \textbf{\#R}: Number of rounds.}%
    \label{tab_best_two_prompt_schemes}%
\end{table}%

\paragraph{Baselines} 
    We consider the state-of-the-art unified fallacy classification model based on T5 \citep{t5} proposed by \citet{t5-multitask-fallacy-detection} as the baseline model since it is the existing most comprehensive SOTA baseline that can be applied to diverse extant fallacy benchmark datasets.
    Baselines in prior works \eg, \cite{dataset-logic} and \cite{elecdebate} can only be applied to one specific fallacy dataset under highly specialized conditions on data annotations, fallacy forms, and additional data features, \etc.
    Due to such limited generalizability and reproducibility, these baselines are currently out of our scope.
    T5 models are implemented in a deployment-efficient unified multitask paradigm, 
    thus serving as a suitable and strong full-shot baseline.
    We follow the instruction-based prompts and hyper-parameter setups used in their original paper (See \Cref{sec:implementation_details}) and replicate three full-shot fine-tuned baseline settings:
    \begin{inparaenum}[(\bgroup\bfseries i\egroup)]
        \item
            \textbf{Single-task} is fine-tuned on each dataset individually.
        \item
            \textbf{Multi-ALR} is fine-tuned on three open-domain datasets \textsc{Argotario}, \textsc{Logic} and \textsc{Reddit}.          
        \item
            \textbf{Multi-ALEP} is fine-tuned on two open-domain datasets \textsc{Argotario}, \textsc{Logic} and two domain-specific datasets {\sc ElecDeb} and {\sc Propaganda}.
    \end{inparaenum}
    We implement each baseline setting with both \textbf{T5-3B} and \textbf{T5-large}. We further apply up-sampling \citet{unifiedskg} techniques to balance the size of multitask training data for better performance during fine-tuning. We report the average results of 3-time repeated experiments for each T5 baseline.

\paragraph{LLMs.} We consider seven recent representative LLMs that have been fine-tuned for powerful instruction following behaviors thus are capable for multi-round chat-based Q\&A: 
    \begin{inparaenum}[(\bgroup\bfseries i\egroup)]
        \item
            \textbf{GPT-4} \citep{gpt4},
        \item
            \textbf{GPT-3.5},           
        \item
            \textbf{Llama3-Chat 8B},
        \item
            \textbf{Qwen2.5-Instruct 14B},
        \item
            \textbf{Qwen2.5-Instruct 7B},
        \item
            \textbf{Mistral-Instruct 7B} \citep{mistral} and 
        \item 
            \textbf{Llama2-Chat 13B}. 
    \end{inparaenum}
    For reproducibility, we report the average results of 3-time repeated experiments for GPTs and the average 5-time repeated experimental results for other small LLMs.

\subsection{Zero-shot Single-round Prompting Schemes vs. Fine-Tuned T5 Baselines}
    We report the results of fine-tuned T5 baselines and LLMs with zero-shot single-round prompting schemes in \Cref{tab_main_results}.
    We have the following observations.
    First, zero-shot single-round prompted GPT-4 can outperform both T5-3B and T5-large fine-tuned baselines on \textsc{Argotario} with the best result of 78.94 and can achieve highly competitive results of 79.10 on \textsc{Reddit} as compared to the best results of 83.20 and 80.42 obtained by T5-3B and T5-large, respectively. 
    Its performances on \textsc{Logic}, \textsc{ElecDeb} and \textsc{Propaganda} are comparable to the best of full-shot T5-large baselines but are still behind the best of T5-3B.
    Except GPT-4, the other LLMs fail to outperform the best fine-tuned T5 baselines across all datasets under a fully zero-shot setting,
    which evidences the difficulty of zero-shot fallacy classification. 
    
    Second, zero-shot prompted LLMs demonstrate stronger generalization abilities than the fine-tuned T5 baselines. 
    As can be seen, LLMs can largely outperform T5 baselines on all the OOD hold-out datasets of \textsc{Mafalda}, \textsc{Covid-19}, \textsc{Reddit}, \textsc{ElecDeb}, and \textsc{Propaganda}.
    Specifically, GPT-4, GPT-3.5 and Qwen2.5 consistently outperform all multitask fine-tuned T5 baselines on four OOD inference scenarios. 
    Llama3 and Mistral can also outperform on two datasets when set as OOD for T5 baselines.
    This validates the poor generalization ability of fine-tuned T5 baselines since they heavily rely on the amount and beneficial diversity of annotated training data thus may struggle in generalizing to OOD scenarios with unseen fallacy classes and discourse types. 
    Besides, when T5 is scaled up, multitask fine-tuning of T5 could be susceptible to the composition of tasks and does not always translate to improvements for all the tasks in training but could result in deteriorated performance compared to single-task training.
    In contrast, resorting to zero-shot prompted LLMs can bypass these limitations.
    Third,
    although zero-shot single-round prompted LLMs show considerably promising performances
    on two open-domain datasets \textsc{Argotario} and \textsc{Reddit}, 
    they are struggling with \textsc{Logic} and \textsc{Mafalda} and the other three domain-specific benchmarks.
    We consider the following reasons:
    \begin{inparaenum}[(\bgroup\bfseries i\egroup)]
        \item
            \textsc{Argotario} and \textsc{Reddit} are two most balanced and comprehensible fallacy datasets that have fallacious discourses with common fallacy types delivered in a relatively intuitive, casual, informal utterances about daily contexts.
            These features may align better with the chat-optimized behaviors of instruction-tuned LLMs. 
        \item
            Although \textsc{Logic} and \textsc{Mafalda} contain open-domain contexts,
            they are difficult because of their large label space (13 and 23 respectively) with edge fallacy types, such as ``{\it Fallacy of Converse}'', ``{\it Doubt Credibility}'', ``{\it Intentional}'', \etc.
            In addition, since \textsc{Logic} is collected from online education websites such as study.com and Quizlet, the noisy contexts in its discourse segments may cause confusion with our instructions.
        \item
            In-domain corpora of news and political speeches are delivered in formal language or specific utterance styles and rely on sufficient contexts for comprehension.
            It could be hard to infer about the fallacy type if the understanding of the content and context is challenged by the truncation of discourses.
            Besides, the fallacious examples in the three in-domain datasets, though may share the same fallacy names with other datasets,
            could have nuanced differences in definitions that deviate from the common ones \citep{t5-multitask-fallacy-detection} and refer to specific language use cases tied to the domain context. 
    \end{inparaenum}

\begin{table}[t]
  \centering
  \resizebox{\linewidth}{!}{
    \begin{tabular}{lrrrr}
    \toprule
    \multirow{2}[2]{*}{\textbf{Model}} & \multicolumn{2}{c}{\textbf{Macro-F1}} & \multicolumn{2}{c}{\textbf{\%Failed}} \\
          & \textbf{\makecell{Infomal\\Def.}} & \textbf{\makecell{Formal\\Def.}} & \textbf{\makecell{Infomal\\Def.}} & \textbf{\makecell{Formal\\Def.}} \\
    \midrule
    GPT-4 & 48.38 & \textbf{49.78} & 0.67  & 0.83 \\
    GPT-3.5 & 31.27 & 27.10  & 2.50   & 4.00 \\
    Qwen2.5-14Bf & 31.11 & \textbf{34.64} & 0.88  & 0.70 \\
    Qwen2.5-7Bf & 30.62 & 29.43 & 5.00     & 2.20 \\
    Llama3-8Bf & 24.43 & \textbf{25.71} & 12.90  & 12.10 \\
    Mistral-7Bf & 22.23 & 18.37 & 3.50   & 3.60 \\
    Llama2-13Bf & 22.09 & 21.30  & 27.70  & 8.10 \\
    \bottomrule
    \end{tabular}%
  }
  \caption{Informal vs. formal fallacy definitions in the single-round prompting scheme on {\sc Mafalda}.}
  \label{tab:formal_vs_informal}%
\end{table}%

\begin{table}[t]
  \centering
  \resizebox{\linewidth}{!}{
    \begin{tabular}{cllll}
    \toprule
    \textbf{Model} & \textbf{Shot} & \multicolumn{1}{c}{\textsc{\textbf{Logic}}} & \multicolumn{1}{c}{\textsc{\textbf{ElecDeb}}} & \multicolumn{1}{c}{\textsc{\textbf{Propaganda}}} \\
    \midrule
    \multirow{4}[2]{*}{GPT-4} & One & \multicolumn{1}{r}{\textbf{54.48}} & \multicolumn{1}{r}{\textbf{44.36}} & \multicolumn{1}{r}{33.18} \\
          & Two & \multicolumn{1}{r}{\textbf{55.14}} & \multicolumn{1}{r}{\textbf{44.64}} & \multicolumn{1}{r}{31.57} \\
          & Zero$^\circ$ & \multicolumn{1}{r}{48.45} & \multicolumn{1}{r}{42.26} & \multicolumn{1}{r}{34.80} \\
          & Zero* & \multicolumn{1}{r}{50.54\textsuperscript{GFA}} & \multicolumn{1}{r}{42.26$^\circ$} & \multicolumn{1}{r}{\textbf{35.37\textsuperscript{CoT}}} \\
    \bottomrule
    \end{tabular}%
    }
    \caption{Few-shot vs. zero-shot results of GPT-4. $^\circ$: Zero-shot single-round results without definitions. *: Best zero-shot results. See \Cref{tab:fs_vs_zs_other_models} for other LLMs.
}
  \label{tab:fs_vs_zs}%
\end{table}%

\begin{figure*}
      \centering
      \includegraphics[width=0.9\linewidth]{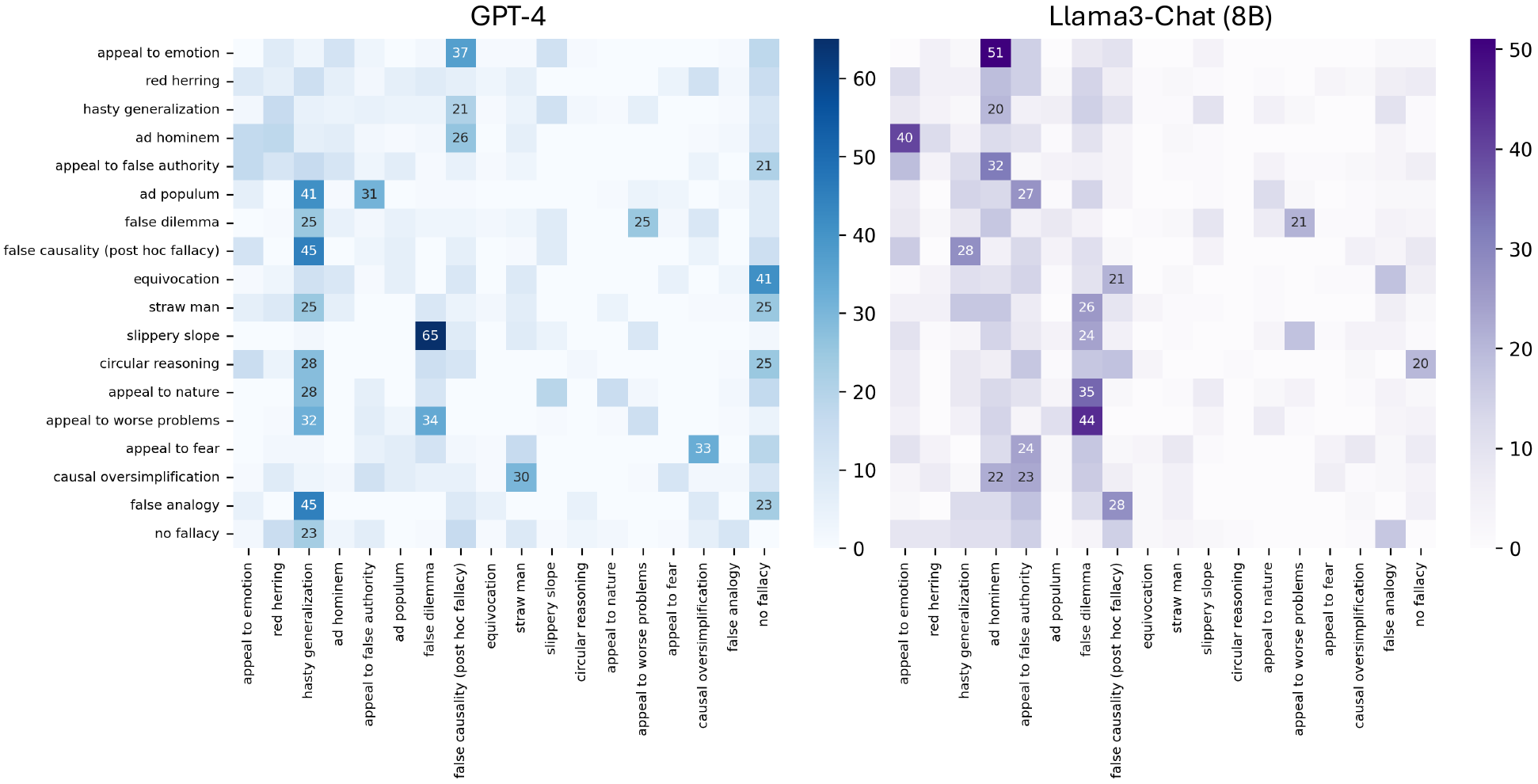}
      \caption{
      Misclassification confusion matrix of common fallacy types given by GPT-4 and Llama3-Chat (8B).
      Rows are the ground truth fallacy types and columns are predicted fallacy types.
      Cell values represent the percentages of row fallacy types that are misclassified as column fallacy types. 
      }
      \vspace{-10pt}
      \label{fig_confusion}
\end{figure*}

\subsection{Single-Round vs. Multi-Round}
    As shown in \Cref{tab_main_results},
    our multi-round prompting schemes are effective in further improving LLMs' zero-shot performances compared to the basic single-round prompts.
    In detail (See \Cref{tab_multiround_gains}), they improve on 83.7\% of the classification experiments compared to the basic single-round prompting schemes,
    with an average improvement across the datasets of 1.63, 2.46, 4.95, 3.67, 7.54, 4.02 and 1.27 points for GPT-4, GPT-3.5, Qwen2.5-Instruct (14B), Qwen2.5-Instruct (7B), Llama3-Chat (8B), Mistral-Instruct (7B) and Llama2-Chat (13B), respectively, 
    indicating the effectiveness of our proposed multi-round prompts, especially for capable small LLMs such as Llama3 and Qwen2.5.
    Multi-round prompted GPT-4 is also further improved on most tasks compared to its best single-round prompting performance and achieves new state-of-the-art scores on \textsc{Argotario}, \textsc{Mafalda} and \textsc{Covid-19}.

\subsection{Multi-Round Prompting Schemes Analysis}
    We present the overall average Macro-F1 ranking of each multi-round prompting scheme in \Cref{tab_scheme_rank}. 
    It shows that three of our proposed multi-round prompting schemes can be superior to the basic single-round prompts.
    General Fallacy Analysis with or without the warm-up round (GFA-W and GFA) ranks within the top three best-performed multi-round schemes,
    indicating that explicit fallacy analysis can benefit LLMs by eliciting their internalized fallacy knowledge to guide the classification process.
    However, all the proposed multi-round prompting schemes that require LLMs to first perform tasks that indirectly address the focal classification problem could be prone to failed classification \ie, 
    predicting a fallacy type that is outside the given fallacy label space or predicting as non-fallacious. 
    This is predicable since it is possible that some data examples do commit more than one type of fallacy or can be interpreted as certain out-of-scope fallacies.
    Having LLMs to make analysis first without limiting them within a specific label space increases the probability that LLMs bias to their first judgements.
    An additional warm-up round that allows LLMs to first reflect on the content and context of input discourses can not only improve the overall classification performance but also help alleviate the failure rate (12.27\% vs. 15.09\%).
    Zero-shot CoT generally ranks as the second-best multi-round prompting scheme, suggesting that CoT's general contribution to improved reasoning in LLMs is also transferable to fallacy classification tasks.
    Besides, Zero-shot CoT has significant advantages over GFA(-W) if taking the average failure rate into account, 
    which makes Zero-shot CoT the potentially most effective and robust multi-round prompting scheme in practice.
    
    Although on average Definition Generation slightly underperforms the single-round With Definitions, it is still better than prompting LLMs with only the label space.
    Human crafted fallacy definitions that could be more detailed and relevant to the focused context, if available, are usually more reliable than LLMs' self-generated definitions. 
    It is notable that both Zero-shot CoT and Definition Generation have lower failure rates, probably because they have fewer chat rounds and can immediately address the given label space at the first place, 
    confining LLMs to more restricted contexts that are less likely to be directed to unlimited or irrelevant association.
    In particular, Definition Generation has the lowest failure rate, indicating that LLMs could be more robust in generating instruction-specified labels when relying on their self-generated contents.
    In contrast, the three-round Premises \& Conclusion that explicitly applies the formal definition of the term ``fallacy'' is not always a robust choice to improve the fallacy classification performance of LLMs.
    It turns out that the formal academic conception cannot be well translated to eliciting LLMs' knowledge and abilities for distinguishing among different fallacy types but mislead LLMs to a deviated focus on checking ``whether the premises entail the conclusion'' rather than recognizing the specific fallacy type.
    (See \Cref{sec:perform_analysis_pec})

    We present the top two best performed prompting schemes for different classification scenarios with respect to data domains and model types in \Cref{tab_best_two_prompt_schemes}.
    This summarization may serve as a general guidance for fallacy classification in practice.
    As suggested by the results, when applying LLMs to open-domain fallacy classification tasks in a zero-shot setting, 
    the inherent pre-trained fallacy knowledge of advanced GPT models could be effective when explicitly prompted for self-generated fallacy definitions or directly referring to human crafted defintions.
    For smaller LLMs, they can be prompted to reflect on the input context before giving a general fallacy analysis to better elicit and leverage their internalized relevant knowledge. 
    When tackling domain-specific fallacy classification tasks, to address the nuances in fallacy definitions for in-domain characteristics, 
    it is more effective to require GPT models to directly refer to the manually crafted definitions, while for small LLMs, they can still benefit from Zero-shot CoT prompting. See \Cref{prompt_scheme_rank_for_LLMs} and \Cref{tab:prompt_scheme_rank_for_datasets} for more detailed rankings and the comparison between best zero-shot results and the guidance-suggested results in \Cref{tab:best_zs_vs_rule}.

\subsection{Informal vs. Formal Definitions}
    We substitute the informal fallacy definitions (\Cref{tab:fallacy_definition}) with formal fallacy definitions (See \Cref{tab:mafalda_formal_defs}) of \textsc{Mafalda} summarized by \citet{benchmark-mafalda} in the single-round prompting scheme. 
    As shown in \Cref{tab:formal_vs_informal}, we can only observe a limited improvement of performance on more capable LLMs of GPT-4, Qwen2.5-Instruct (14B) and Llama3-Chat (8B) when providing them with formal definitions of fallacies. 
    The results of the other LLMs remain comparable with or are slightly worse than that with informal definitions.
    The failure rates of the two prompting schemes are not significantly different.
    While formal definitions that incorporate some symbolic representations could be less ambiguous,
    they are not necessarily easier for LLMs to comprehend and relate to the exact occurrences of fallacies in actual discourses.  

\subsection{Few-shot vs. Zero-shot}
    To verify whether LLMs' weak performance on some datasets (\ie, \textsc{Logic}, \textsc{ElecDeb} and \textsc{Propaganda}) could have been resulted from the challenging fully zero-shot prompt setting, we conduct N-way few-shot experiments that randomly present LLMs with 1-shot or 2-shot examples of each fallacy classes from the holdout data splits. 
    Experiment results in \Cref{tab:fs_vs_zs} show that only GPT-4's performance can be further improved on {\sc Logic} and \textsc{ElecDeb}. The rest of few-shot results by other LLMs all fail to surpass the previous best-performed zero-shot prompting results,
    suggesting the limitation of LLMs' in-context learning abilities in handling hard and in-domain fallacy classification tasks.

\subsection{Error Analysis}
    To view how LLMs perform on different fallacy types, we present the confusion matrices of GPT-4 and Llama3-Chat (8B) on each dataset in \Cref{sec:confusion_matrix_details}. 
    To further analyze classification errors, we aggregate the experimental results of GPT-4 and Llama3 on each dataset under their top three best performed zero-shot prompting schemes and select a total of 18 common fallacy types (including {\it No Fallacy}) that occur in more than two datasets to present the confusion matrix in misclassification percentage as shown in \Cref{fig_confusion}. 
    It is notable that GPT-4 is generally more inclined to classify fallacy examples to be non-fallacious compared to Llama3. 
    As revealed in \Cref{fig:cm_argotario} (a) and \Cref{fig:cm_mafalda_propaganda_logic} (e), GPT-4 can better recall actual {\it No Fallacy} examples but severely over predict most examples in \textsc{Covid-19} to be non-fallacious as shown in \Cref{fig:cm_covid_reddit_elecdeb} (b). 
    In addition, some fallacy types frequently confused by LLMs reflect the inherent limitation in the natural language definitions of informal fallacies, which may overlap with each other with inevitable ambiguity.
    For example, GPT-4 tends to classify many fallacy types as {\it Hasty Generalization}. 
    We find that GPT-4's self-generated definition for {\it Hasty Generalization} that ``{\it A conclusion based on insufficient or biased evidence; rushing to a conclusion before you have all relevant facts}.'' can be applied to interpret many fallacies \eg, {\it Ad Populum}, {\it Post Hoc Fallacy} and {\it False Analogy}.
    We find similar explanations towards both GPT-4's and Llama3's misclassifations of {\it Appeal to Worse Problems}, {\it Slippery Slope} as {\it False Dilemma} and {\it Ad Populum} as {\it Appeal to False Authority}. 
    As for Llama3, we find that it tends to overpredict the fallacy types of {\it False Dilemma}, {\it Ad Hominem} and {\it Appeal to False Authority}.
    Interestingly, as shown in the corpus of U.S. presidential debate \textsc{ElecDeb} in \Cref{fig:cm_covid_reddit_elecdeb} (d), Llama3 predicts more {\it Appeal to Emotion}, {\it Ad Hominem} and {\it Appeal to False Authority}, which are more basic and superficial fallacy types, while GPT-4 predicts more {\it Slippery Slope} and {\it Post Hoc Fallacy} that are fallacy conceptions more related to causal reasoning. This further suggests that the task of fallacy classification requires advanced reasoning abilities.
    We further analyze the potential reason for the relatively poor performance of GPT-based models under the zero-shot multi-round prompting scheme of Premises \& Conclusion in \Cref{sec:perform_analysis_pec}.

\section{Conclusion}
    In this paper, we explore LLMs' performance on zero-shot fallacy classification.
    We propose both single-round and multi-round prompting schemes to fully elicit LLMs' fallacy classification ability.
    Through extensive experiments on benchmark datasets,
    we demonstrate that LLMs with a zero-shot single-round prompting scheme can outperform or achieve highly comparable performances with SOTA full-shot fine-tuned T5 baselines on some open-domain benchmark datasets and can generally achieve sub-optimal performances on hard domain-specific datasets.
    Besides, our multi-round prompting schemes can effectively enhance the performances, especially for small LLMs.
    The performance gaps between zero-shot prompted LLMs and full-shot fine-tuned baselines are acceptable, indicating LLMs' potential for further improvements for this task.
    Our detailed analysis moreover highlights the future research for zero-shot fallacy classification.

\section*{Limitations}
    We consider the following limitations for future work.
    First, we only focus on single-label fallacy classification.
    For instance, we only use the dominant types in the dataset \textsc{Mafalda} as the ground-truth labels.
    We may later explore the more complex multi-label fallacy classification task.
    Second, since we only conduct few-shot experiments on instruction-tuned LLMs that are optimized for chat-based applications, 
    it is yet to be explored whether vanilla LLMs could better leverage their in-context learning abilities for the few-shot classification. 

\bibliography{custom}
\clearpage
\appendix
\def\eg{\emph{e.g.}} 
\def\etc{\emph{etc.}}
\onecolumn
\section{Implementation Details}
\label{sec:implementation_details}
We fine-tune all T5 baselines and run inference with small LLMs (Qwen2.5-Instruct 14B, Qwen2.5-Instruct 7B, Llama3-Chat 8B, Llama2-Chat 13B and Mistral-Instruct 7B) on two RTX A6000 48GB GPUs.

We basically follow the hyperparameter setting used in \citet{t5-multitask-fallacy-detection} for the single/multi-task training of T5-large and T5-3B. We train all T5 models for 5 epochs and select the best performed checkpoints on validation sets as the final models for inference. We use a constant 1e-4 learning rate with warm-up and set batch sizes and gradient accumulation steps with respect to training data size accordingly. We report the results given by the Adafactor optimizer as it performs better than AdamW in our setting. The maximum input length is 1024 so contexts beyond the limit are truncated. The maximum generation length is 64 and the number of beams is 1. The generated output is compared with the ground truth based on strict string matching. Fine-tuning the T5-3B for 5 epochs takes 4 to 6 hours on average, varying with the size of the dataset.

We follow the default generation hyperparameter configurations for inference with all LLMs \eg, sampling is used for generation, with temperature of 0.75 (0.6 for Llama3), top p of 0.9 and top k of 50 \etc.

Due to the lack of published preprocessed dataset splits in previous work,
we have to adopt different dataset setups tailored to our experiments to ensure that our test splits have acceptable sizes of sufficient statistical power to test for LLMs’ performance. 
In particular, the complete dataset of \textsc{Covid-19} has become proprietary by the time we start this work. We only have access to a small piece of sample dataset that is publicly available. The results of this dataset may vary if more data are included.
We exclude some fallacy classes in \textsc{ElecDeb} and \textsc{Propaganda} following \cite{benchmark-mafalda} to exclude propaganda-like techniques that do not align well with the definition of ``fallacy''.
As a result of the above changes in setting, our T5 baseline results under similar conditions do not strictly reproduce the results of previous works.

We experimented with two versions of the three-round prompting schemes Premises \& Conclusion to thoroughly examine the effects of prompting LLMs with the formal definition of ``fallacy'' \citep{benchmark-mafalda}. We aim to explore whether explicitly prompting LLMs to reason based on this standard definition could enhance LLMs' understanding of fallacies and translate to improvement on the classification performances. The two versions differ in the position of the definition text and the instruction utterance to cope with any potential ambiguity. Based on our analysis of results derived from both of these two versions, we provide our interpretations of the reasons why this multi-round prompting scheme underperforms.

\section{Experiment Results}
\label{sec:experiment_result_details}
\subsection{Zero-shot Single-round Prompting Results} 
We report detailed experimental results (including accuracy scores and failure rates) of single-round prompting schemes with (\textbf{w/}) or without (\textbf{w/o}) definitions for each LLM on each dataset compared to the best results of the T5-3B and T5-large baselines.

\begin{minipage}[c]{0.48\textwidth}
 \centering
  \resizebox{\linewidth}{!}{
    \begin{tabular}{cllrrr}
    \toprule
    \textbf{Setting} & \textbf{Model} & \textbf{Def.} & \textbf{Macro-F1} & \textbf{Acc.} & \textbf{\%Failed} \\
    \midrule
    \multirow{2}[2]{*}{Full} & T5-3B Multi-ALR & w/    & 72.38 & 73.08 & 0.00 \\
          & T5-large Multi-ALR & w/    & 65.65 & 67.63 & 0.00 \\
    \midrule
    \multirow{14}[14]{*}{\begin{sideways}Zero-shot Single-round\end{sideways}} & \multirow{2}[2]{*}{GPT-4} & w/    & 78.94 & 78.96 & 0.64 \\
          &       & w/o   & 78.62 & 78.53 & 1.39 \\
\cmidrule{2-6}          & \multirow{2}[2]{*}{GPT-3.5} & w/o   & 63.59 & 63.59 & 1.54 \\
          &       & w/    & 61.72 & 62.50 & 5.82 \\
\cmidrule{2-6}          & \multirow{2}[2]{*}{Qwen2.5-14Bf} & w/    & 68.19 & 67.95 & 2.88 \\
          &       & w/o   & 67.84 & 68.40 & 2.05 \\
\cmidrule{2-6}          & \multirow{2}[2]{*}{Qwen2.5-7Bf} & w/    & 61.38 & 63.40 & 0.00 \\
          &       & w/o   & 59.59 & 62.18 & 0.06 \\
\cmidrule{2-6}          & \multirow{2}[2]{*}{Llama3-8Bf} & w/    & 48.87 & 53.53 & 1.02 \\
          &       & w/o   & 46.57 & 52.24 & 1.34 \\
\cmidrule{2-6}          & \multirow{2}[2]{*}{Mistral-7Bf} & w/o   & 57.04 & 60.13 & 2.88 \\
          &       & w/    & 50.92 & 56.60 & 1.99 \\
\cmidrule{2-6}          & \multirow{2}[2]{*}{Llama2-13Bf} & w/    & 50.20 & 55.00 & 0.13 \\
          &       & w/o   & 47.39 & 53.27 & 0.51 \\
    \bottomrule
    \end{tabular}%
    }
    \captionof{table}{Single-round results on \textsc{Argotario}}
  \label{tab:b-argotario}%
\end{minipage}
\begin{minipage}[c]{0.48\textwidth}
  \centering
  \resizebox{\linewidth}{!}{
    \begin{tabular}{cllrrr}
    \toprule
    \textbf{Setting} & \textbf{Model} & \textbf{Def.} & \textbf{Macro-F1} & \textbf{Acc.} & \textbf{\%Failed} \\
    \midrule
    \multirow{2}[2]{*}{\begin{sideways}Full\end{sideways}} & T5-3B Single-task & w/    & 64.95  & 70.89  & 0.33  \\
          & T5-large Multi-ALR & w/    & 59.48  & 65.22  & 0.00  \\
    \midrule
    \multirow{14}[14]{*}{\begin{sideways}Zero-shot Single-round\end{sideways}} & \multirow{2}[2]{*}{GPT-4} & w/    & 50.43  & 59.78  & 5.89  \\
          &   & w/o   & 48.45  & 58.11  & 9.44  \\
\cmidrule{2-6}          & \multirow{2}[2]{*}{GPT-3.5} & w/    & 39.65  & 49.44  & 11.45  \\
          &   & w/o   & 38.60  & 47.78  & 11.22  \\
\cmidrule{2-6}          & \multirow{2}[2]{*}{Qwen2.5-14Bf} & w/o   & 41.82  & 52.07  & 4.27  \\
          &   & w/    & 38.36  & 50.20  & 4.00  \\
\cmidrule{2-6}          & \multirow{2}[2]{*}{Qwen2.5-7Bf} & w/o   & 35.48  & 38.53  & 4.47  \\
          &   & w/    & 30.99  & 39.27  & 3.13  \\
\cmidrule{2-6}          & \multirow{2}[2]{*}{Llama3-8Bf} & w/o   & 27.45  & 40.67  & 3.27  \\
          &   & w/    & 27.34  & 39.80  & 1.67  \\
\cmidrule{2-6}          & \multirow{2}[2]{*}{Mistral-7Bf} & w/o   & 28.99  & 40.47  & 8.40  \\
          &   & w/    & 26.07  & 37.41  & 5.67  \\
\cmidrule{2-6}          & \multirow{2}[2]{*}{Llama2-13Bf} & w/    & 25.11  & 36.27  & 1.93  \\
          &   & w/o   & 24.06  & 32.73  & 6.60  \\
    \bottomrule
    \end{tabular}%
    }
  \captionof{table}{Single-round results on \textsc{Logic}}
  \label{tab:b-logic}%
\end{minipage}%

\begin{minipage}[c]{0.48\textwidth}
  \centering
  \resizebox{\linewidth}{!}{
    \begin{tabular}{cllrrr}
    \toprule
    \textbf{Setting} & \textbf{Model} & \textbf{Def.} & \textbf{Macro-F1} & \textbf{Acc.} & \textbf{\%Failed} \\
    \midrule
    \multirow{2}[2]{*}{\begin{sideways}Full\end{sideways}} & T5-3B Single-task & w/    & 83.20  & 83.30  & 0.06  \\
          & T5-large Multi-ALR & w/    & 80.42  & 80.64  & 0.13  \\
    \midrule
    \multirow{14}[14]{*}{\begin{sideways}Zero-shot Single-round\end{sideways}} & \multirow{2}[2]{*}{GPT-4} & w/    & 79.10  & 79.73  & 2.34  \\
          &  & w/o   & 77.70  & 78.49  & 3.70  \\
\cmidrule{2-6}          & \multirow{2}[2]{*}{GPT-3.5} & w/o   & 70.42  & 70.76  & 2.40  \\
          &  & w/    & 68.90  & 67.90  & 2.66  \\
\cmidrule{2-6}          & \multirow{2}[2]{*}{Qwen2.5-14Bf} & w/o   & 67.72  & 68.13  & 3.17  \\
          &  & w/    & 66.01  & 66.86  & 7.02  \\
\cmidrule{2-6}          & \multirow{2}[2]{*}{Qwen2.5-7Bf} & w/    & 58.58  & 59.61  & 0.62  \\
          &  & w/o   & 58.14  & 59.96  & 2.42  \\
\cmidrule{2-6}          & \multirow{2}[2]{*}{Llama3-8Bf} & w/    & 49.41  & 52.09  & 2.53  \\
          &  & w/o   & 45.21  & 48.26  & 4.72  \\
\cmidrule{2-6}          & \multirow{2}[2]{*}{Mistral-7Bf} & w/o   & 46.89  & 46.20  & 8.38  \\
          &  & w/    & 45.92  & 45.03  & 9.01  \\
\cmidrule{2-6}          & \multirow{2}[2]{*}{Llama2-13Bf} & w/o   & 34.15  & 38.21  & 21.52  \\
          &  & w/    & 34.14  & 38.87  & 15.36  \\
    \bottomrule
    \end{tabular}%
    }
    \captionof{table}{Single-round results on \textsc{Reddit}}
  \label{tab:b-reddit}%
\end{minipage}
\begin{minipage}[c]{0.48\textwidth}
  \centering
  \resizebox{\linewidth}{!}{
    \begin{tabular}{clllrr}
    \toprule
    \textbf{Setting} & \textbf{Model} & \textbf{Def.} & \textbf{Macro-F1} & \textbf{Acc.} & \textbf{\%Failed} \\
    \midrule
    \multirow{2}[2]{*}{\begin{sideways}Full\end{sideways}} & T5-3B Single-task & w/    & 62.37  & 79.78  & 0.00  \\
          & T5-large Multi-ALEP & w/    & 56.15  & 71.78  & 0.00  \\
    \midrule
    \multirow{14}[14]{*}{\begin{sideways}Zero-shot Single-round\end{sideways}} & \multirow{2}[2]{*}{GPT-4} & w/o   & 42.26  & 45.56  & 6.71  \\
          &  & w/    & 41.93  & 45.78  & 9.96  \\
\cmidrule{2-6}          & \multirow{2}[2]{*}{GPT-3.5} & w/    & 41.01  & 43.33  & 3.47  \\
          &  & w/o   & 37.10  & 37.11  & 1.30  \\
\cmidrule{2-6}          & \multirow{2}[2]{*}{Qwen2.5-14Bf} & w/    & 37.28  & 41.20  & 8.70  \\
          &  & w/o   & 35.76  & 43.87  & 12.99  \\
\cmidrule{2-6}          & \multirow{2}[2]{*}{Qwen2.5-7Bf} & w/    & 43.34  & 55.33  & 4.42  \\
          &  & w/o   & 42.28  & 53.33  & 1.82  \\
\cmidrule{2-6}          & \multirow{2}[2]{*}{Llama3-8Bf} & w/o   & 39.36  & 63.60  & 0.26  \\
          &  & w/    & 36.00  & 53.60  & 0.00  \\
\cmidrule{2-6}          & \multirow{2}[2]{*}{Mistral-7Bf} & w/o   & 33.23  & 40.93  & 4.80  \\
          &  & w/    & 28.01  & 36.83  & 4.06  \\
\cmidrule{2-6}          & \multirow{2}[2]{*}{Llama2-13Bf} & w/    & 35.57  & 55.07  & 0.78  \\
          &  & w/o   & 32.69  & 59.87  & 3.12  \\
    \bottomrule
    \end{tabular}%
    }
  \captionof{table}{Single-round results on {\sc ElecDeb}}
  \label{tab:b-elecdeb}%
\end{minipage}

\begin{minipage}[c]{0.48\textwidth}
  \centering
   \resizebox{\linewidth}{!}{
    \begin{tabular}{cllrrr}
    \toprule
    \textbf{Setting} & \textbf{Model} & \textbf{Def.} & \textbf{Macro-F1} & \textbf{Acc.} & \textbf{\%Failed} \\
    \midrule
    \multirow{2}[2]{*}{\begin{sideways}Full\end{sideways}} & T5-3B Multi-ALEP & w/    & 43.33  & 76.73  & 0.25  \\
          & T5-large Multi-ALEP & w/    & 39.75  & 76.48  & 0.00  \\
    \midrule
    \multirow{14}[14]{*}{\begin{sideways}Zero-shot Single-round\end{sideways}} & \multirow{2}[2]{*}{GPT-4} & w/o   & 34.80  & 58.49  & 2.39  \\
          &  & w/    & 33.45  & 56.10  & 2.39  \\
\cmidrule{2-6}          & \multirow{2}[2]{*}{GPT-3.5} & w/    & 22.39  & 38.11  & 3.90  \\
          &  & w/o   & 21.84  & 34.59  & 4.28  \\
\cmidrule{2-6}          & \multirow{2}[2]{*}{Qwen2.5-14Bf} & w/    & 21.89  & 41.13  & 6.61  \\
          &  & w/o   & 16.04  & 29.81  & 16.23  \\
\cmidrule{2-6}          & \multirow{2}[2]{*}{Qwen2.5-7Bf} & w/    & 16.03  & 33.66  & 1.28  \\
          &  & w/o   & 15.66  & 24.53  & 0.00  \\
\cmidrule{2-6}          & \multirow{2}[2]{*}{Llama3-8Bf} & w/    & 17.30  & 28.68  & 0.15  \\
          &  & w/o   & 16.03  & 33.73  & 1.58  \\
\cmidrule{2-6}          & \multirow{2}[2]{*}{Mistral-7Bf} & w/o   & 16.89  & 24.60  & 10.72  \\
          &  & w/    & 11.75  & 19.93  & 14.19  \\
\cmidrule{2-6}          & \multirow{2}[2]{*}{Llama2-13Bf} & w/    & 10.61  & 11.32  & 3.09  \\
          &  & w/o   & 4.41  & 8.91  & 9.06  \\
    \bottomrule
    \end{tabular}%
    }
  \captionof{table}{Single-round results on {\sc Propaganda}}
  \label{tab:b-propaganda}%
\end{minipage}%
\begin{minipage}[c]{0.48\textwidth}
  \centering
  \resizebox{\linewidth}{!}{
    \begin{tabular}{cllrrr}
    \toprule
    \textbf{Setting} & \textbf{Model} & \textbf{Def.} & \textbf{Macro-F1} & \textbf{Acc.} & \textbf{\%Failed} \\
    \midrule
    \multirow{2}[2]{*}{\begin{sideways}Full\end{sideways}} & T5-3B Multi-ALEP & w/    & 35.60  & 31.83  & 17.50  \\
          & T5-large Multi-ALEP & w/    & 25.60  & 27.33  & 7.17  \\
    \midrule
    \multirow{14}[14]{*}{\begin{sideways}Zero-shot Single-round\end{sideways}} & \multirow{2}[2]{*}{GPT-4} & w/o   & 48.74  & 63.50  & 1.50  \\
          &  & w/    & 48.38  & 65.50  & 0.67  \\
\cmidrule{2-6}          & \multirow{2}[2]{*}{GPT-3.5} & w/    & 31.27  & 44.67  & 2.50  \\
          &  & w/o   & 28.97  & 42.00  & 2.33  \\
\cmidrule{2-6}          & \multirow{2}[2]{*}{Qwen2.5-14Bf} & w/o   & 33.03  & 50.70  & 0.70  \\
          &  & w/    & 31.11  & 49.00  & 0.88  \\
\cmidrule{2-6}          & \multirow{2}[2]{*}{Qwen2.5-7Bf} & w/o   & 31.27  & 41.50  & 2.80  \\
          &  & w/    & 30.62  & 37.00  & 5.00  \\
\cmidrule{2-6}          & \multirow{2}[2]{*}{Llama3-8Bf} & w/o   & 24.85  & 29.10  & 10.80  \\
          &  & w/    & 24.43  & 27.40  & 12.90  \\
\cmidrule{2-6}          & \multirow{2}[2]{*}{Mistral-7Bf} & w/o   & 23.44  & 41.50  & 4.80  \\
          &  & w/    & 22.23  & 36.30  & 3.50  \\
\cmidrule{2-6}          & \multirow{2}[2]{*}{Llama2-13Bf} & w/    & 22.09  & 28.30  & 27.70  \\
          &  & w/o   & 17.60  & 27.40  & 32.00  \\
    \bottomrule
    \end{tabular}%
    }
  \captionof{table}{Single-round results on {\sc Mafalda}}
  \label{tab:b-mafalda}%
\end{minipage}%

\begin{minipage}[c]{0.48\textwidth}
  \centering
  \resizebox{\linewidth}{!}{
    \begin{tabular}{cllrrr}
    \toprule
    \textbf{Setting} & \textbf{Model} & \textbf{Def.} & \textbf{Macro-F1} & \textbf{Acc.} & \textbf{\%Failed} \\
    \midrule
    \multirow{2}[2]{*}{\begin{sideways}Full\end{sideways}} & T5-3B Multi-ALEP & w/    & 14.59  & 19.05  & 0.65  \\
          & T5-large Multi-ALEP & w/    & 14.08  & 18.40  & 0.65  \\
    \midrule
    \multirow{14}[14]{*}{\begin{sideways}Zero-shot Single-round\end{sideways}} & \multirow{2}[2]{*}{GPT-4} & w/    & 20.47  & 34.63  & 0.22  \\
          &  & w/o   & 16.74  & 31.17  & 0.00  \\
\cmidrule{2-6}          & \multirow{2}[2]{*}{GPT-3.5} & w/    & 17.45  & 20.35  & 1.95  \\
          &  & w/o   & 14.88  & 20.78  & 2.17  \\
\cmidrule{2-6}          & \multirow{2}[2]{*}{Qwen2.5-14Bf} & w/    & 17.60  & 27.27  & 1.56  \\
          &  & w/o   & 15.97  & 26.10  & 1.30  \\
\cmidrule{2-6}          & \multirow{2}[2]{*}{Qwen2.5-7Bf} & w/o   & 15.19  & 30.52  & 1.30  \\
          &  & w/    & 13.48  & 27.66  & 1.69  \\
\cmidrule{2-6}          & \multirow{2}[2]{*}{Llama3-8Bf} & w/o   & 11.00  & 18.70  & 0.26  \\
          &  & w/    & 9.67  & 16.88  & 0.00  \\
\cmidrule{2-6}          & \multirow{2}[2]{*}{Mistral-7Bf} & w/    & 14.69  & 18.83  & 2.27  \\
          &  & w/o   & 10.23  & 19.35  & 4.29  \\
\cmidrule{2-6}          & \multirow{2}[2]{*}{Llama2-13Bf} & w/    & 14.15  & 15.32  & 1.69  \\
          &  & w/o   & 13.09  & 22.47  & 18.05  \\
    \bottomrule
    \end{tabular}%
    }
  \captionof{table}{Single-round results on {\sc Covid-19}}
  \label{tab:b-covid}%
\end{minipage}%
\begin{minipage}[c]{0.49\textwidth}
  \centering
  \resizebox{\linewidth}{!}{
    \begin{tabular}{cllrrr}
    \toprule
    \textbf{Setting} & \textbf{Model} & \textbf{Scheme} & \textbf{Macro-F1} & \textbf{Acc.} & \textbf{\%Failed} \\
    \midrule
    \multirow{2}[2]{*}{\begin{sideways}Full\end{sideways}} & T5-3B Multi-ALR & WD    & 72.38  & 73.08  & 0.00  \\
          & T5-large Multi-ALR & WD    & 65.65  & 67.63  & 0.00  \\
    \midrule
    \multirow{14}[14]{*}{\begin{sideways}Zero-shot Multi-round\end{sideways}} & \multirow{2}[2]{*}{GPT-4} & DG    & 79.87  & 80.24  & 0.32  \\
          &  & CoT   & 76.85  & 76.60  & 0.19  \\
\cmidrule{2-6}          & \multirow{2}[2]{*}{GPT-3.5} & GFA   & 68.40  & 67.84  & 2.24  \\
          &  & DG    & 66.77  & 68.19  & 0.40  \\
\cmidrule{2-6}          & \multirow{2}[2]{*}{Qwen2.5-14Bf} & CoT   & 68.87  & 70.19  & 0.00  \\
          &  & GFA   & 67.63  & 66.92  & 3.01  \\
\cmidrule{2-6}          & \multirow{2}[2]{*}{Qwen2.5-7Bf} & DG    & 60.20  & 63.20  & 0.26  \\
          &  & GFA-W & 57.71  & 59.36  & 2.05  \\
\cmidrule{2-6}          & \multirow{2}[2]{*}{Llama3-8Bf} & P\&C$^{2}$ & 61.39  & 63.59  & 17.57  \\
          &  & GFA-W & 59.25  & 61.22  & 16.86  \\
\cmidrule{2-6}          & \multirow{2}[2]{*}{Mistral-7Bf} & GFA-W & 57.26  & 58.97  & 7.24  \\
          &  & GFA   & 55.84  & 56.99  & 12.88  \\
\cmidrule{2-6}          & \multirow{2}[2]{*}{Llama2-13Bf} & GFA   & 48.79  & 54.30  & 5.90  \\
          &  & DG    & 45.27  & 50.51  & 0.13  \\
    \bottomrule
    \end{tabular}%
    }
\captionof{table}{Multi-round results on \sc{Argotario}}
\label{tab:argotario-b2}%
\end{minipage}

\subsection{Zero-shot Multi-round Prompting Results}
We report the detailed experiment results of the best two multi-round prompting schemes for each LLM on each dataset compared to the best results of the T5-3B and T5-large baselines.
Notations for multi-round prompting schemes: \textbf{DG} for Definition Generation. \textbf{GFA} for General Fallacy Analysis. \textbf{GFA-W} for General Fallacy Analysis with Warm up. \textbf{P\&C} for Premises \& Conclusion. \textbf{CoT} for Zero-shot CoT. 

\begin{minipage}[c]{0.48\textwidth}
  \centering
  \resizebox{\linewidth}{!}{
    \begin{tabular}{cllrrr}
    \toprule
    \textbf{Setting} & \textbf{Model} & \textbf{Scheme} & \textbf{Macro-F1} & \textbf{Acc.} & \textbf{\%Failed} \\
    \midrule
    \multirow{2}[2]{*}{\begin{sideways}Full\end{sideways}} & T5-3B Single-task & WD    & 64.95  & 70.89  & 0.33  \\
          & T5-large Multi-ALR & WD    & 59.48  & 65.22  & 0.00  \\
    \midrule
    \multirow{14}[14]{*}{\begin{sideways}Zero-shot Multi-round\end{sideways}} & \multirow{2}[2]{*}{GPT-4} & GFA   & 50.54  & 58.78  & 8.22  \\
          &  & DG    & 50.03  & 59.67  & 9.11  \\
\cmidrule{2-6}          & \multirow{2}[2]{*}{GPT-3.5} & GFA   & 41.11  & 51.45  & 17.11  \\
          &  & GFA-W & 39.09  & 50.89  & 15.33  \\
\cmidrule{2-6}          & \multirow{2}[2]{*}{Qwen2.5-14Bf} & CoT   & 45.89  & 54.00  & 6.25  \\
          &  & GFA   & 42.50  & 50.67  & 19.53  \\
\cmidrule{2-6}          & \multirow{2}[2]{*}{Qwen2.5-7Bf} & CoT   & 40.09  & 47.27  & 10.87  \\
          &  & GFA-W & 36.22  & 46.13  & 14.07  \\
\cmidrule{2-6}          & \multirow{2}[2]{*}{Llama3-8Bf} & GFA-W & 35.66  & 47.27  & 16.40  \\
          &  & GFA   & 35.18  & 45.40  & 15.80  \\
\cmidrule{2-6}          & \multirow{2}[2]{*}{Mistral-7Bf} & GFA-W & 31.43  & 44.13  & 16.80  \\
          &  & GFA   & 31.42  & 44.33  & 15.47  \\
\cmidrule{2-6}          & \multirow{2}[2]{*}{Llama2-13Bf} & GFA   & 28.85  & 39.80  & 28.00  \\
          &  & GFA-W & 28.24  & 39.80  & 21.87  \\
    \bottomrule
    \end{tabular}%
    }
  \captionof{table}{Multi-round results on \sc{Logic}}
  \label{tab:logic-b2}%
\end{minipage}%
\begin{minipage}[c]{0.49\textwidth}
  \centering
  \resizebox{\linewidth}{!}{
    \begin{tabular}{cllrrr}
    \toprule
    \textbf{Setting} & \textbf{Model} & \textbf{Scheme} & \textbf{Macro-F1} & \textbf{Acc.} & \textbf{\%Failed} \\
    \midrule
    \multirow{2}[2]{*}{\begin{sideways}Full\end{sideways}} & T5-3B Single-task & WD    & 62.37  & 79.78  & 0.00  \\
          & T5-large Multi-ALEP & WD    & 56.15  & 71.78  & 0.00  \\
    \midrule
    \multirow{14}[14]{*}{\begin{sideways}Zero-shot Multi-round\end{sideways}} & \multirow{2}[2]{*}{GPT-4} & DG    & 41.25  & 43.55  & 8.01  \\
          &  & GFA-W & 36.27  & 40.22  & 25.76  \\
\cmidrule{2-6}          & \multirow{2}[2]{*}{GPT-3.5} & DG    & 37.77  & 40.22  & 2.17  \\
          &  & CoT   & 34.55  & 35.67  & 3.41  \\
\cmidrule{2-6}          & \multirow{2}[2]{*}{Qwen2.5-14Bf} & GFA-W & 34.09  & 42.27  & 25.20  \\
          &  & GFA   & 33.82  & 42.40  & 28.57  \\
\cmidrule{2-6}          & \multirow{2}[2]{*}{Qwen2.5-7Bf} & DG    & 44.55  & 55.87  & 0.91  \\
          &  & GFA-W & 31.20  & 35.86  & 17.01  \\
\cmidrule{2-6}          & \multirow{2}[2]{*}{Llama3-8Bf} & CoT   & 40.81  & 57.07  & 2.60  \\
          &  & DG    & 36.09  & 56.67  & 0.26  \\
\cmidrule{2-6}          & \multirow{2}[2]{*}{Mistral-7Bf} & DG    & 32.91  & 42.40  & 4.54  \\
          &  & CoT   & 27.82  & 42.53  & 23.77  \\
\cmidrule{2-6}          & \multirow{2}[2]{*}{Llama2-13Bf} & DG    & 36.37  & 49.07  & 0.78  \\
          &  & GFA-W & 29.93  & 35.07  & 17.01  \\
    \bottomrule
    \end{tabular}%
    }
  \captionof{table}{Multi-round results on {\sc ElecDeb}}
  \label{tab:elecdebate-b2}%
\end{minipage}%

\begin{minipage}[c]{0.48\textwidth}
  \centering  
  \resizebox{\linewidth}{!}{
    \begin{tabular}{cllrrr}
    \toprule
    \textbf{Setting} & \textbf{Model} & \textbf{Scheme} & \textbf{Macro-F1} & \textbf{Acc.} & \textbf{\%Failed} \\
    \midrule
    \multirow{2}[2]{*}{\begin{sideways}Full\end{sideways}} & T5-3B Multi-ALEP & WD    & 43.33  & 76.73  & 0.25  \\
          & T5-large Multi-ALEP & WD    & 39.75  & 76.48  & 0.00  \\
    \midrule
    \multirow{14}[14]{*}{\begin{sideways}Zero-shot Multi-round\end{sideways}} & \multirow{2}[2]{*}{GPT-4} & CoT   & 35.37  & 52.08  & 5.66  \\
          &  & DG    & 33.23  & 57.23  & 4.28  \\
\cmidrule{2-6}          & \multirow{2}[2]{*}{GPT-3.5} & CoT   & 26.67  & 37.55  & 3.87  \\
          &  & DG    & 23.98  & 40.13  & 1.89  \\
\cmidrule{2-6}          & \multirow{2}[2]{*}{Qwen2.5-14Bf} & CoT   & 26.60  & 46.51  & 10.38  \\
          &  & GFA-W & 22.54  & 32.23  & 13.21  \\
\cmidrule{2-6}          & \multirow{2}[2]{*}{Qwen2.5-7Bf} & GFA   & 19.22  & 30.79  & 13.05  \\
          &  & GFA-W & 16.88  & 24.60  & 13.96  \\
\cmidrule{2-6}          & \multirow{2}[2]{*}{Llama3-8Bf} & CoT   & 21.35  & 35.32  & 5.89  \\
          &  & GFA-W & 18.95  & 35.40  & 26.34  \\
\cmidrule{2-6}          & \multirow{2}[2]{*}{Mistral-7Bf} & CoT   & 20.41  & 40.53  & 6.42  \\
          &  & DG    & 15.68  & 26.34  & 7.25  \\
\cmidrule{2-6}          & \multirow{2}[2]{*}{Llama2-13Bf} & GFA-W & 11.11  & 12.53  & 18.94  \\
          &  & GFA   & 10.13  & 13.36  & 35.70  \\
    \bottomrule
    \end{tabular}%
    }
  \captionof{table}{Multi-round results on {\sc Propaganda}}
  \label{tab:propaganda-b2}%
\end{minipage}%
\begin{minipage}[c]{0.48\textwidth}
  \centering
  \resizebox{\linewidth}{!}{
    \begin{tabular}{cllrrr}
    \toprule
    \textbf{Setting} & \textbf{Model} & \textbf{Scheme} & \textbf{Macro-F1} & \textbf{Acc.} & \textbf{\%Failed} \\
    \midrule
    \multirow{2}[2]{*}{\begin{sideways}Full\end{sideways}} & T5-3B Multi-ALEP & WD    & 35.60  & 31.83  & 17.50  \\
          & T5-large Multi-ALEP & WD    & 25.60  & 27.33  & 7.17  \\
    \midrule
    \multirow{14}[14]{*}{\begin{sideways}Zero-shot Multi-round\end{sideways}} & \multirow{2}[2]{*}{GPT-4} & CoT   & 52.86  & 58.50  & 1.83  \\
          &  & GFA-W & 48.55  & 57.50  & 0.50  \\
\cmidrule{2-6}          & \multirow{2}[2]{*}{GPT-3.5} & GFA-W & 40.73  & 52.33  & 2.50  \\
          &  & GFA   & 36.85  & 44.67  & 4.17  \\
\cmidrule{2-6}          & \multirow{2}[2]{*}{Qwen2.5-14Bf} & P\&C$^{2}$ & 45.94  & 58.00  & 0.25  \\
          &  & CoT   & 43.17  & 52.25  & 0.50  \\
\cmidrule{2-6}          & \multirow{2}[2]{*}{Qwen2.5-7Bf} & P\&C$^{1}$ & 35.37  & 52.50  & 0.30  \\
          &  & GFA   & 34.98  & 44.40  & 0.30  \\
\cmidrule{2-6}          & \multirow{2}[2]{*}{Llama3-8Bf} & P\&C$^{1}$ & 34.18  & 41.33  & 9.00  \\
          &  & GFA-W & 33.71  & 39.90  & 14.10  \\
\cmidrule{2-6}          & \multirow{2}[2]{*}{Mistral-7Bf} & P\&C$^{1}$ & 29.08  & 49.50  & 1.62  \\
          &  & GFA-W & 28.55  & 45.20  & 2.60  \\
\cmidrule{2-6}          & \multirow{2}[2]{*}{Llama2-13Bf} & GFA-W & 15.68  & 39.70  & 1.00  \\
          &  & CoT   & 15.50  & 17.70  & 8.70  \\
    \bottomrule
    \end{tabular}%
    }
  \captionof{table}{Multi-round results on {\sc Mafalda}}
  \label{tab:mafalda-b2}%
\end{minipage}%

\begin{minipage}[c]{0.48\textwidth}
  \centering
  \resizebox{\linewidth}{!}{
    \begin{tabular}{cllrrr}
    \toprule
    \textbf{Setting} & \textbf{Model} & \textbf{Scheme} & \textbf{Macro-F1} & \textbf{Acc.} & \textbf{\%Failed} \\
    \midrule
    \multirow{2}[2]{*}{\begin{sideways}Full\end{sideways}} & T5-3B Single-task & WD    & 83.20  & 83.30  & 0.06  \\
          & T5-large Multi-ALR & WD    & 80.42  & 80.64  & 0.13  \\
    \midrule
    \multirow{14}[14]{*}{\begin{sideways}Zero-shot Multi-round\end{sideways}} & \multirow{2}[2]{*}{GPT-4} & CoT   & 81.11  & 81.68  & 3.70  \\
          &  & DG    & 79.34  & 80.25  & 2.08  \\
\cmidrule{2-6}          & \multirow{2}[2]{*}{GPT-3.5} & DG    & 71.08  & 71.41  & 3.70  \\
          &  & CoT   & 68.64  & 68.88  & 2.14  \\
\cmidrule{2-6}          & \multirow{2}[2]{*}{Qwen2.5-14Bf} & CoT   & 77.08  & 77.29  & 1.95  \\
          &  & GFA-W & 71.08  & 71.15  & 7.29  \\
\cmidrule{2-6}          & \multirow{2}[2]{*}{Qwen2.5-7Bf} & P\&C$^{2}$ & 64.58  & 65.81  & 2.49  \\
          &  & CoT   & 64.08  & 65.58  & 4.52  \\
\cmidrule{2-6}          & \multirow{2}[2]{*}{Llama3-8Bf} & P\&C$^{1}$ & 57.83  & 58.64  & 19.77  \\
          &  & GFA-W & 57.45  & 57.66  & 20.00  \\
\cmidrule{2-6}          & \multirow{2}[2]{*}{Mistral-7Bf} & CoT   & 59.70  & 60.82  & 12.52  \\
          &  & DG    & 49.27  & 48.62  & 4.44  \\
\cmidrule{2-6}          & \multirow{2}[2]{*}{Llama2-13Bf} & GFA   & 45.82  & 46.70  & 22.96  \\
          &  & P\&C$^{2}$ & 44.49  & 45.11  & 11.70  \\
    \bottomrule
    \end{tabular}%
    }
  \captionof{table}{Multi-round results on {\sc Reddit}}
  \label{tab:reddit-b2}%
\end{minipage}%
\begin{minipage}[c]{0.49\textwidth}
  \centering
  \resizebox{\linewidth}{!}{
    \begin{tabular}{cllrrr}
    \toprule
    \textbf{Setting} & \textbf{Model} & \textbf{Scheme} & \textbf{Macro-F1} & \textbf{Acc.} & \textbf{\%Failed} \\
    \midrule
    \multirow{2}[2]{*}{\begin{sideways}Full\end{sideways}} & T5-3B Multi-ALEP & WD    & 14.59  & 19.05  & 0.65  \\
          & T5-large Multi-ALEP & WD    & 14.08  & 18.40  & 0.65  \\
    \midrule
    \multirow{14}[14]{*}{\begin{sideways}Zero-shot Multi-round\end{sideways}} & \multirow{2}[2]{*}{GPT-4} & CoT   & 25.18  & 33.12  & 0.00  \\
          &  & GFA-W & 23.59  & 40.69  & 1.73  \\
\cmidrule{2-6}          & \multirow{2}[2]{*}{GPT-3.5} & GFA   & 17.24  & 30.74  & 5.41  \\
          &  & GFA-W & 16.04  & 29.22  & 2.38  \\
\cmidrule{2-6}          & \multirow{2}[2]{*}{Qwen2.5-14Bf} & GFA   & 23.73  & 30.00  & 4.67  \\
          &  & CoT   & 23.31  & 34.09  & 0.00  \\
\cmidrule{2-6}          & \multirow{2}[2]{*}{Qwen2.5-7Bf} & GFA   & 22.88  & 33.77  & 0.39  \\
          &  & CoT   & 19.71  & 31.69  & 0.00  \\
\cmidrule{2-6}          & \multirow{2}[2]{*}{Llama3-8Bf} & CoT   & 19.83  & 28.44  & 0.13  \\
          &  & GFA   & 16.20  & 19.22  & 30.00  \\
\cmidrule{2-6}          & \multirow{2}[2]{*}{Mistral-7Bf} & CoT   & 18.53  & 26.62  & 0.78  \\
          &  & P\&C$^{1}$ & 16.41  & 27.92  & 5.58  \\
\cmidrule{2-6}          & \multirow{2}[2]{*}{Llama2-13Bf} & DG    & 14.16  & 30.26  & 0.00  \\
          &  & P\&C$^{1}$ & 12.76  & 29.22  & 3.64  \\
    \bottomrule
    \end{tabular}%
    }
  \captionof{table}{Multi-round results on {\sc Covid-19}}
  \label{tab:covid-b2}%
\end{minipage}%

\subsection{Prompting Scheme Rankings}
\begin{table}[!ht]
  \centering
  \setlength{\tabcolsep}{30pt}
  \resizebox{\linewidth}{!}{
    \begin{tabular}{cllrrr}
    \toprule
    \textbf{Data Domain} & \textbf{LLM} & \textbf{Scheme} & \textbf{\#R} & \textbf{Rank} & \textbf{\%Failed} \\
    \midrule
    \multirow{12}[12]{*}{\textbf{Open}} & \multirow{2}[2]{*}{GPT-4} & Zero-shot CoT & 2     & 2.50  & 3.43 \\
          &       & With Definitions & 1     & 2.75  & 2.38 \\
\cmidrule{2-6}          & \multirow{2}[2]{*}{GPT-3.5} & General Fallacy Analysis & 2     & 2.50  & 9.40 \\
          &       & Definition Generation & 2     & 2.75  & 3.51 \\
\cmidrule{2-6}          & \multicolumn{1}{l}{\multirow{2}[2]{*}{Qwen2.5-7/14Bf}} & Zero-shot CoT & 2     & 2.25  & 4.13 \\
          &       & General Fallacy Analysis with Warm Up & 3     & 3.50  & 7.06 \\
\cmidrule{2-6}          & \multirow{2}[2]{*}{Llama3-8Bf} & Premises \& Conclusion & 3     & 2.62  & 15.58 \\
          &       & General Fallacy Analysis with Warm Up & 3     & 2.80  & 16.81 \\
\cmidrule{2-6}          & \multirow{2}[2]{*}{Mistral-7Bf} & General Fallacy Analysis with Warm Up & 3     & 2.25  & 12.76 \\
          &       & Zero-shot CoT & 2     & 2.67  & 9.13 \\
\cmidrule{2-6}          & \multirow{2}[2]{*}{Llama2-13Bf} & With Definitions & 1     & 4.83  & 7.84 \\
          &       & Definition Generation & 2     & 5.50  & 1.43 \\
    \midrule
    \multirow{12}[12]{*}{\textbf{Specific}} & \multirow{2}[2]{*}{GPT-4} & Zero-shot CoT & 2     & 2.33  & 13.36 \\
          &       & With Definitions & 1     & 3.33  & 4.19 \\
\cmidrule{2-6}          & \multirow{2}[2]{*}{GPT-3.5} & With Definitions & 1     & 1.67  & 3.11 \\
          &       & Definition Generation & 2     & 3.00  & 2.22 \\
\cmidrule{2-6}          & \multicolumn{1}{l}{\multirow{2}[2]{*}{Qwen2.5-7/14Bf}} & General Fallacy Analysis & 2     & 2.83  & 14.22 \\
          &       & General Fallacy Analysis with Warm Up & 3     & 3.00  & 12.10 \\
\cmidrule{2-6}          & \multirow{2}[2]{*}{Llama3-8Bf} & Zero-shot CoT & 2     & 1.00  & 2.87 \\
          &       & General Fallacy Analysis with Warm Up & 3     & 3.67  & 18.74 \\
\cmidrule{2-6}          & \multirow{2}[2]{*}{Mistral-7Bf} & Zero-shot CoT & 2     & 2.00  & 10.32 \\
          &       & Without Definitions & 1     & 3.67  & 6.60 \\
\cmidrule{2-6}          & \multirow{2}[2]{*}{Llama2-13Bf} & With Definitions & 1     & 2.00  & 1.85 \\
          &       & Definition Generation & 2     & 2.67  & 0.76 \\
    \bottomrule
    \end{tabular}%
  }
  \captionof{table}{The best two prompting schemes for different LLMs and data domains based on average Macro-F1 rankings. \textbf{\#R}: Number of rounds. \textbf{Rank}: Average ranking on Macro-F1, the lower the better.
}
  \label{prompt_scheme_rank_for_LLMs}%
\end{table}

\begin{table}[!ht]
  \centering
   \setlength{\tabcolsep}{45pt}
   \resizebox{\linewidth}{!}{
    \begin{tabular}{cllrrr}
    \toprule
    \textbf{Domain} & \textbf{Task} & \textbf{Scheme} & \textbf{\#R} & \textbf{Rank} & \textbf{\%Failed} \\
    \midrule
    \multirow{8}[8]{*}{\textbf{Open}} & \multirow{2}[2]{*}{\textsc{\textbf{Argotario}}} & Definition Generation & 2     & 5.40  & 0.30 \\
          &       & With Definitions & 1     & 5.60  & 1.44 \\
\cmidrule{2-6}          & \multirow{2}[2]{*}{\textsc{\textbf{Logic}}} & General Fallacy Analysis & 2     & 1.71  & 17.60 \\
          &       & General Fallacy Analysis with Warm Up & 3     & 2.57  & 15.13 \\
\cmidrule{2-6}          & \multirow{2}[2]{*}{\textsc{\textbf{Mafalda}}} & General Fallacy Analysis with Warm Up & 3     & 2.57  & 3.21 \\
          &       & General Fallacy Analysis & 2     & 3.67  & 5.12 \\
\cmidrule{2-6}          & \multirow{2}[2]{*}{\textsc{\textbf{Reddit}}} & Zero-shot CoT & 2     & 3.00  & 4.98 \\
          &       & General Fallacy Analysis with Warm Up & 3     & 3.71  & 15.03 \\
    \midrule
    \multirow{6}[6]{*}{\textbf{Specific}} & \multirow{2}[2]{*}{\textsc{\textbf{Covid-19}}} & General Fallacy Analysis & 2     & 2.86  & 6.99 \\
          &       & Zero-shot CoT & 2     & 3.00  & 0.18 \\
\cmidrule{2-6}          & \multirow{2}[2]{*}{\textsc{\textbf{ElecDeb}}} & Without Definitions & 1     & 2.14  & 4.43 \\
          &       & With Definitions & 1     & 2.14  & 4.48 \\
\cmidrule{2-6}          & \multirow{2}[2]{*}{\textsc{\textbf{Propaganda}}} & Zero-shot CoT & 2     & 2.43  & 6.87 \\
          &       & General Fallacy Analysis with Warm Up & 3     & 3.00  & 19.14 \\
    \bottomrule
    \end{tabular}}%
    \caption{The best two prompting schemes for each benchmark dataset based on average Macro-F1 rankings. \textbf{\#R}: Number of rounds. \textbf{Rank}: Average ranking on Macro-F1, the lower the better.
}
  \label{tab:prompt_scheme_rank_for_datasets}%
\end{table}%

\newpage
\subsection{Few-shot vs. Zero-shot Results} 
\begin{table}[!ht]
  \centering
  \setlength{\tabcolsep}{40pt}
  \resizebox{0.95\linewidth}{!}{
    \begin{tabular}{llrrr}
    \toprule
    \textbf{Model} & \textbf{Shot} & \multicolumn{1}{c}{\textsc{\textbf{Logic}}} & \multicolumn{1}{c}{\textsc{\textbf{ElecDeb}}} & \multicolumn{1}{c}{\textsc{\textbf{Propaganda}}} \\
    \midrule
    \multirow{4}[2]{*}{GPT-3.5} & One   & 37.13 & 31.41 & 12.23 \\
          & Two   & 34.55 & 24.07 & 12.08 \\
          & Zero$^\circ$  & 38.6  & 37.1  & 21.84 \\
          & Zero* & \textbf{41.11\textsuperscript{GFA}} & \textbf{41.01\textsuperscript{WD}} & \textbf{26.67\textsuperscript{CoT}} \\
    \midrule
    \multirow{4}[2]{*}{Qwen2.5-14Bf} & One   & 43.08 & 27.91 & 20.54 \\
          & Two   & 44.47 & 31.07 & 20.73 \\
          & Zero$^\circ$  & 41.82 & 35.76 & 16.04 \\
          & Zero* & \textbf{45.89\textsuperscript{CoT}} & \textbf{37.28\textsuperscript{WD}} & \textbf{26.6\textsuperscript{CoT}} \\
    \midrule
    \multirow{4}[2]{*}{Qwen2.5-7Bf} & One   & 33.22 & 35.53 & 13.67 \\
          & Two   & 28.62 & 31.69 & 15.58 \\
          & Zero$^\circ$  & 35.48 & 42.28 & 15.66 \\
          & Zero* & \textbf{40.09\textsuperscript{CoT}} & \textbf{44.55\textsuperscript{DG}} & \textbf{18.95\textsuperscript{GFA}} \\
    \midrule
    \multirow{4}[2]{*}{Llama3-8Bf} & One   & 22.94 & 27.2  & 7.55 \\
          & Two   & 24.64 & 24.16 & 6.32 \\
          & Zero$^\circ$  & 27.45 & 39.36 & 16.03 \\
          & Zero* & \textbf{35.66\textsuperscript{GFA-W}} & \textbf{40.81\textsuperscript{CoT}} & \textbf{21.35\textsuperscript{CoT}} \\
    \midrule
    \multirow{4}[2]{*}{Mistral-7Bf} & One   & 30.08 & 25.32 & 11.61 \\
          & Two   & 27.04 & 24.95 & 8.55 \\
          & Zero$^\circ$  & 28.99 & 33.23 & 16.89 \\
          & Zero* & \textbf{31.43\textsuperscript{GFA-W}} & \textbf{33.23\textsuperscript{WoD}} & \textbf{20.41\textsuperscript{CoT}} \\
    \midrule
    \multirow{4}[2]{*}{Llama2-13Bf} & One   & 1.76  & 3.73  & 1.47 \\
          & Two   & 1.74  & 7.22  & 3.15 \\
          & Zero$^\circ$  & 24.06 & 32.69 & 4.41 \\
          & Zero* & \textbf{28.85\textsuperscript{GFA}} & \textbf{36.37\textsuperscript{DG}} & \textbf{11.11\textsuperscript{GFA-W}} \\
    \bottomrule
    \end{tabular}%
    }
    \caption{Few-shot vs. zero-shot results. $^\circ$: Zero-shot single-round results without definitions. *: Best zero-shot results.
}
  \label{tab:fs_vs_zs_other_models}%
\end{table}%

\begin{table*}[!ht]
  \centering
  \resizebox{0.95\linewidth}{!}{
    \begin{tabular}{lrrrrrrrr}
    \toprule
    \textbf{Model} & \textsc{\textbf{Argotario}} & \textsc{\textbf{Logic}} & \textsc{\textbf{Reddit}} & \textsc{\textbf{ElecDebate}} & \textsc{\textbf{Propaganda}} & \textsc{\textbf{Mafalda}} & \textsc{\textbf{Covid-19}} & \textbf{Avg.} \\
    \midrule
    GPT-4 & 0.93  & 0.11  & 2.01  & -1.01 & 0.57  & 4.12  & 4.71  & \textbf{1.63} \\
    GPT-3.5 & 4.81  & 1.46  & 0.66  & -3.24 & 4.28  & 9.46  & -0.21 & \textbf{2.46} \\
    Qwen2.5-Instruct 14B & 0.68  & 4.07  & 9.36  & -3.19 & 4.71  & 12.91 & 6.13  & \textbf{4.95} \\
    Qwen2.5-Instruct 7B & -1.18  & 4.61  & 6.00     & 1.21  & 3.19  & 4.10   & 7.69  & \textbf{3.66} \\
    Llama3-Chat 8B & 12.52 & 8.21  & 8.42  & 1.45  & 4.05  & 9.33  & 8.83  & \textbf{7.54} \\
    Mistral-Instruct 7B & 0.22   & 2.44  & 12.81 & -0.32 & 3.52  & 5.64  & 3.84  & \textbf{4.02} \\
    Llama2-Chat 13B & -1.41 & 3.74  & 11.67 & 0.80   & 0.50   & -6.41 & 0.01  & \textbf{1.27} \\
    \bottomrule
    \end{tabular}%
    }
    \caption{
        Performance improvements of the best performed multi-round prompting schemes compared to the best performed single-round schemes. The average performance gains across datasets for each LLM are in \textbf{bold}.
    }
  \label{tab_multiround_gains}%
\end{table*}%

\begin{table*}[!ht]
  \centering
  \setlength{\tabcolsep}{10pt}
  \resizebox{0.95\linewidth}{!}{
    \begin{tabular}{lrrrrrrr}
    \toprule
    \textbf{Model} & \multicolumn{1}{c}{\textbf{\textsc{Argotario}}} & \multicolumn{1}{c}{\textbf{\textsc{Logic}}} & \multicolumn{1}{c}{\textbf{\textsc{Reddit}}} & \multicolumn{1}{c}{\textbf{\textsc{ElecDeb}}} & \multicolumn{1}{c}{\textbf{\textsc{Propaganda}}} & \multicolumn{1}{c}{\textbf{\textsc{Mafalda}}} & \multicolumn{1}{c}{\textbf{\textsc{Covid-19}}} \\
    \midrule
    \multirow{2}[1]{*}{GPT-4} & \textbf{79.87\textsuperscript{DG}} & \textbf{50.54\textsuperscript{GFA}} & \textbf{81.11\textsuperscript{CoT}} & \textbf{41.25\textsuperscript{DG}} & \textbf{35.37\textsuperscript{CoT}} & \textbf{52.86\textsuperscript{CoT}} & \textbf{25.18\textsuperscript{CoT}} \\
          & 79.87\textsuperscript{DG} & 50.03\textsuperscript{DG} & 79.34\textsuperscript{DG} & 41.93\textsuperscript{WD} & 33.45\textsuperscript{WD} & 45.85\textsuperscript{DG} & 20.47\textsuperscript{WD} \\
    \midrule
    \multirow{2}[0]{*}{GPT-3.5} & \textbf{68.4\textsuperscript{GFA}} & \textbf{41.11\textsuperscript{GFA}} & \textbf{71.08\textsuperscript{DG}} & \textbf{37.77\textsuperscript{DG}} & \textbf{26.67\textsuperscript{CoT}} & \textbf{40.73\textsuperscript{GFA-W}} & \textbf{17.24\textsuperscript{GFA}} \\
          & 66.77\textsuperscript{DG} & 37.2\textsuperscript{DG} & 71.08\textsuperscript{DG} & 41.01\textsuperscript{WD} & 22.39\textsuperscript{WD} & 35.25\textsuperscript{DG} & 17.45\textsuperscript{WD} \\
    \midrule
    \multirow{2}[0]{*}{Qwen2.5-14Bf} & \textbf{68.87\textsuperscript{CoT}} & \textbf{45.89\textsuperscript{CoT}} & \textbf{77.08\textsuperscript{CoT}} & \textbf{34.09\textsuperscript{GFA-W}} & \textbf{26.6\textsuperscript{CoT}} & \textbf{45.94\textsuperscript{P\&C$^{2}$}} & \textbf{23.73\textsuperscript{GFA}} \\
          & 66.91\textsuperscript{GFA-W} & 41.64\textsuperscript{GFA-W} & 71.08\textsuperscript{GFA-W} & 32.31\textsuperscript{CoT} & 26.6\textsuperscript{CoT} & 40.53\textsuperscript{GFA-W} & 23.31\textsuperscript{CoT} \\
    \midrule
    \multirow{2}[0]{*}{Qwen2.5-7Bf} & \textbf{60.2\textsuperscript{DG}} & \textbf{40.09\textsuperscript{CoT}} & \textbf{64.58\textsuperscript{P\&C$^{2}$}} & \textbf{44.55\textsuperscript{DG}} & \textbf{19.22\textsuperscript{GFA}} & \textbf{35.37\textsuperscript{P\&C$^{1}$}} & \textbf{22.88\textsuperscript{GFA}} \\
          & 57.71\textsuperscript{GFA-W} & 36.22\textsuperscript{GFA-W} & 61.69\textsuperscript{GFA-W} & 30.89\textsuperscript{CoT} & 15.57\textsuperscript{CoT} & 33.2\textsuperscript{GFA-W} & 19.71\textsuperscript{CoT} \\
    \midrule
    \multirow{2}[0]{*}{Llama3-8Bf} & \textbf{61.39\textsuperscript{P\&C$^{2}$}} & \textbf{35.66\textsuperscript{GFA-W}} & \textbf{57.83\textsuperscript{P\&C$^{1}$}} & \textbf{40.81\textsuperscript{CoT}} & \textbf{21.35\textsuperscript{CoT}} & \textbf{34.18\textsuperscript{P\&C$^{1}$}} & \textbf{19.83\textsuperscript{CoT}} \\
          & 59.25\textsuperscript{GFA-W} & 35.66\textsuperscript{GFA-W} & 57.45\textsuperscript{GFA-W} & 40.81\textsuperscript{CoT} & 21.35\textsuperscript{CoT} & 33.71\textsuperscript{GFA-W} & 19.83\textsuperscript{CoT} \\
    \midrule
    \multirow{2}[0]{*}{Mistral-7Bf} & \textbf{57.26\textsuperscript{GFA-W}} & \textbf{31.43\textsuperscript{GFA-W}} & \textbf{59.7\textsuperscript{CoT}} & \textbf{32.91\textsuperscript{DG}} & \textbf{20.41\textsuperscript{CoT}} & \textbf{29.08\textsuperscript{P\&C$^{1}$}} & \textbf{18.53\textsuperscript{CoT}} \\
          & 57.26\textsuperscript{GFA-W} & 31.43\textsuperscript{GFA-W} & 44.84\textsuperscript{GFA-W} & 27.82\textsuperscript{CoT} & 20.41\textsuperscript{CoT} & 28.55\textsuperscript{GFA-W} & 18.53\textsuperscript{CoT} \\
    \midrule
    \multirow{2}[1]{*}{Llama2-13Bf} & \textbf{48.79\textsuperscript{GFA}} & \textbf{28.85\textsuperscript{GFA}} & \textbf{45.82\textsuperscript{GFA}} & \textbf{36.37\textsuperscript{DG}} & \textbf{11.11\textsuperscript{GFA-W}} & \textbf{15.68\textsuperscript{GFA-W}} & \textbf{14.16\textsuperscript{DG}} \\
          & 44.42\textsuperscript{GFA-W} & 28.24\textsuperscript{GFA-W} & 42.68\textsuperscript{GFA-W} & 27.44\textsuperscript{CoT} & 7.72\textsuperscript{CoT} & 15.68\textsuperscript{GFA-W} & 8.82\textsuperscript{CoT} \\
    \bottomrule
    \end{tabular}%
    }
    \caption{
        Best zero-shot results vs. results given by the general domain-model guidance based on prompt scheme rankings. The best zero-shot results are in \textbf{bold}.
    }
  \label{tab:best_zs_vs_rule}%
\end{table*}%

\onecolumn
\begin{table}[htbp]
  \centering
  \resizebox{\linewidth}{!}{
    \begin{NiceTabular}{l|ccccccc}
    \toprule
    \textbf{Fallacy Type} & \textsc{\textbf{Argotario}} & \textsc{\textbf{Logic}} & \textsc{\textbf{Reddit}} & \textsc{\textbf{Mafalda}} & \textsc{\textbf{ElecDeb}} & \textsc{\textbf{Propaganda}} & \textsc{\textbf{Covid-19}} \\
    \midrule
    Appeal to False Authority & 40    &       & 64    & 6     & 22    & 10    & 10 \\
    Ad Hominem & 42    & 41    &       & 7     & 21    &       &  \\
    Appeal to Emotion & 69    & 23    &       & 5     & 96    &       &  \\
    Red Herring & 37    & 24    &       &       &       & 3     & 9 \\
    Hasty Generalization & 53    &       & 60    & 20    &       &       & 7 \\
    Ad Populum &       & 30    & 59    & 12    &       & 1     &  \\
    False Dilemma &       & 12    & 63    & 6     &       & 10    &  \\
    False Causality (Post Hoc Fallacy) &       & 18    &       & 12    & 6     &       & 19 \\
    Equivocation &       & 5     &       & 4     &       & 1     & 7 \\
    Straw Man &       & 21    &       & 2     &       & 2     & 5 \\
    Slippery Slope &       &       & 69    & 9     & 5     &       &  \\
    Circular Reasoning &       & 19    &       & 8     &       &       &  \\
    Appeal to Nature &       &       & 63    & 9     &       &       &  \\
    Appeal to Worse Problems &       &       & 72    & 5     &       &       &  \\
    Doubt Credibility &       & 17    &       &       &       & 47    &  \\
    Fallacy of Converse (Affirming the Consequent) &       & 14    &       &       &       &       &  \\
    Appeal to Fear &       &       &       & 5     &       & 10    &  \\
    Causal Oversimplification &       &       &       & 16    &       & 19    &  \\
    Faulty Generalization &       & 61    &       &       &       &       &  \\
    Intentional (Intentionally Wrong Argument) &       & 15    &       &       &       &       &  \\
    Appeal to Tradition &       &       & 63    &       &       &       &  \\
    Appeal to Ridicule &       &       &       & 7     &       &       &  \\
    Fallacy of Division &       &       &       & 2     &       &       &  \\
    False Analogy &       &       &       & 5     &       &       & 8 \\
    Guilt by Association &       &       &       & 3     &       &       &  \\
    Tu Quoque &       &       &       & 3     &       &       &  \\
    Flag-Waving &       &       &       &       &       & 21    &  \\
    Name-calling &       &       &       &       &       & 120   &  \\
    Reductio Ad Hitlerum &       &       &       &       &       & 5     &  \\
    Whataboutism &       &       &       &       &       & 5     &  \\
    Cherry Picking &       &       &       &       &       &       & 13 \\
    Evading the Burden of Proof &       &       &       &       &       &       & 14 \\
    \bottomrule
    \end{NiceTabular}%
    }
    \caption{Distribution of included fallacies on each dataset.}
  \label{tab:dataset_details}%
\end{table}%

{\small
  \centering
    \begin{longtblr}[
caption = {Fallacy definitions},
  label = {tab:fallacy_definition}
                    ]{
  cells={valign=m},
  colspec = {|X[l,.20\linewidth]| X[l]|},}
           \hline
    \SetCell[c=1]{c}{\textbf{Fallacy}} & \SetCell[c=1]{c}{\textbf{Definition}} \\
    \hline
    Hasty Generalization & a fallacy when someone makes generalizations based on incomplete observations that cannot represent or generalize to other situations if other relevant factors are taken into account. \\
\cline{1-2}    Red Herring & a fallacy when someone introduces irrelevant or confusing information in arguments to diverge attention from the main topic being discussed to irrelevant issues. \\
\cline{1-2}   \SetCell[r=2]{l}{Circular Reasoning} & ({\sc Logic})a fallacy when the conclusion of an argument is a restatement of the assumption, or the argument assumes the very thing it is trying to prove. \\
          & ({\sc Mafalda})This fallacy occurs when an argument assumes the very thing it is trying to prove, resulting in a circular and logically invalid argument. \\
\cline{1-2}   \SetCell[r=2]{l}{Causal Oversimplification} & ({\sc Propaganda})a fallacy that assumes only a single cause or reason for an issue when there were actually multiple ones. \\
          & ({\sc Mafalda})This fallacy occurs when a complex issue is reduced to a single cause and effect, oversimplifying the actual relationships between events or factors. \\
\cline{1-2}    Doubt Credibility & a propaganda technique that attacks or questions the credibility of someone or something in order to discredit the opponent's argument. \\
\cline{1-2}   \SetCell[r=2]{l}{Appeal to False Authority} & ({\sc Argotario, Elecdeb, COVID-19,Propaganda}) a fallacy when someone attempts to argue or persuade by referring to the opinions or statements of a questionable authority who lacks sufficient credibility in the discussed matter because the authority's expertise may be inadequate/irrelevant or the authority is attributed a statement that has been tweaked. \\
          & ({\sc Mafalda})This fallacy occurs when an argument relies on the opinion or endorsement of an authority figure who may not have relevant expertise or whose expertise is questionable. When applicable, a scientific consensus is not an appeal to authority. \\
\cline{1-2}    {False Causality (Post Hoc Fallacy)} & a fallacy that incorrectly assumes that one event causes another solely based on the observation of a temporal order or correlation that one event came before the other, rather than a proven causal relationship. \\
\cline{1-2}\SetCell[r=3]{l}{Ad Hominem} & ({\sc Argotario})a fallacy when someone attacks the others' characters or motives instead of addressing the substance of their arguments. \\
        & ({\sc Elecdeb, Logic})a fallacy when the arguer directly attacks the opponent's characters, positions or motives instead of addressing the substance of their arguments. \\
        & ({\sc Mafalda})This fallacy involves attacking a person's character or motives instead of addressing the substance of their argument. \\
\cline{1-2}    Appeal to Emotion & a fallacy when the arguer attempts to argue or persuade by using emotive language to arouse non-rational sentiments within the intended audience. \\
\cline{1-2}   \SetCell[r=2]{l}{Ad Populum} & ({\sc Logic,Reddit,Propaganda})a fallacy when the arguer claims that an idea or action is true, valid, correct, or better simply because it is popular or widely accepted by the majority of people, or something is unreal, invalid or bad because few people believe in it. \\
          & ({\sc Mafalda})This fallacy involves claiming that an idea or action is valid because it is popular or widely accepted.  \\
\cline{1-2}   \SetCell[r=2]{l}{False Dilemma} & ({\sc Logic,Reddit,Propaganda})a fallacy that forces a conclusion by presenting or implying an incomplete list (usually two) of options or sides, even though in fact there are more that can be chosen from. \\
          & ({\sc Mafalda})This fallacy occurs when only two options are presented in an argument, even though more options may exist. \\
\cline{1-2}    \SetCell[r=3]{l}{Equivocation} & ({\sc Logic})a fallacy when an argument uses ambiguous language or changing the meaning of a term from time to time in an attempt to confuse or obfuscate. \\
          & (Propaganda, COVID-19)a fallacy that uses confused, unclear or intenationally vague languages to disguise the argument. \\
          & ({\sc Mafalda})This fallacy involves using ambiguous language or changing the meaning of a term within an argument, leading to confusion and false conclusions. \\
\cline{1-2}    \SetCell[r=2]{l}{Straw Man} & ({\sc Logic, Propaganda, COVID-19})a fallacy when the arguer substitutes an opponent's argument with a distorted, exaggerated or misinterpreted version to make it more easily to be attacked or discredited. \\
          & ({\sc Mafalda})This fallacy involves misrepresenting an opponent's argument, making it easier to attack and discredit. \\
\cline{1-2}
\SetCell[r=2]{l}{Slippery Slope} & ({\sc Reddit, Elecdeb})a fallacy when the arguer claims that a small or insignificant action step will inevitably lead to a chain of events that result in significant negative consequences, while the connection between such events or steps is unwarranted or improbable. \\
          & ({\sc Mafalda})This fallacy occurs when it is claimed that a small step will inevitably lead to a chain of events, resulting in a significant negative outcome. \\
\cline{1-2}    Appeal to Nature & a fallacy that occurs when something is assumed to be good or desirable simply because it is natural, while its unnatural counterpart is assumed to be bad or undesirable. \\
\cline{1-2}    Appeal to Worse Problems & This fallacy involves dismissing an issue or problem by claiming that there are more important issues to deal with, instead of addressing the argument at hand. This fallacy is also known as the ``relative privation'' fallacy. \\
\cline{1-2}    {Fallacy of Converse (Affirming the Consequent)} & a fallacy that takes a true conditional statement under certain assumptions but invalidly infers its converse even though the conversed statement may not be true under the same assumptions. \\
\cline{1-2}    Appeal to Fear or Prejudice & a fallacy that uses fear, not based on evidence or reason, as the primary motivator to get others to accept an idea, proposition, or conclusion. \\
\cline{1-2}    Faulty Generalization & a fallacy when someone makes generalizations based on incomplete/partial observations that cannot represent or generalize to all the possible situations (or the whole population) if other relevant factors are taken into account. \\
\cline{1-2}    \SetCell[r=1]{l}{Intentional (Intentionally Wrong Argument)} & a fallacy that uses some intentional or subconscious actions or choices to incorrectly support an argument. \\
\cline{1-2}   \SetCell[r=2]{l}{Appeal to Tradition} & ({\sc Reddit})a fallacy that involves arguing that something should continue to be done a certain way because it has always been done that way, rather than evaluating its merits. \\
          & ({\sc Mafalda})This fallacy involves arguing that something should continue to be done a certain way because it has always been done that way, rather than evaluating its merits. \\
\cline{1-2}    Appeal to Ridicule & This fallacy occurs when an opponent's argument is portrayed as absurd or ridiculous with the intention of discrediting it. \\
\cline{1-2}    Fallacy of Division & This fallacy involves assuming that if something is true for a whole, it must also be true of all or some of its parts. \\
\cline{1-2}    False Analogy & a fallacy that involves making an analogy between two elements based on superficial resemblance. \\
\cline{1-2}    Guilt by association & This fallacy involves discrediting an idea or person based on their association with another person, group, or idea that is viewed negatively. \\
\cline{1-2}    Tu Quoque & This fallacy occurs when someone's argument is dismissed because they are accused of acting inconsistently with their claim, rather than addressing the argument itself. \\
\cline{1-2}    Flag-Waving & a fallacy that plays on strong feeling (e.g. patriotism or nationalism) for a group/nation/country to justify or promote an action or idea. \\
\cline{1-2}    Name-calling & a fallacy that labels the object of campaign as either something the target audience fears, hates, finds undesirable or loves, praises. \\
\cline{1-2}    Reductio Ad Hitlerum & a fallacy that persuades an audience to disapprove an action or idea by suggesting that the idea is popular with groups hated in contempt (e.g., Hitler or the Nazi party) by the target audience. \\
\cline{1-2}    Whataboutism & a fallacy that responses to an accusation of wrongdoing by claiming that an offense committed by another is similar or worse, attempting to discredit the allegation. \\
\cline{1-2}    Cherry Picking & a fallacy that chooses among competing evidence only for those that support a given position or argument, ignoring or dismissing findings which do not support it. \\
\cline{1-2}    Evading the Burden of Proof & a fallacy that persuades an argument without any support of evidence as if it was self-evident \\
    \hline
    \end{longtblr}%
}

{\small
  \centering
    \begin{longtblr}[
caption = {Formal fallacy definitions for \textsc{Mafalda}},
  label = {tab:mafalda_formal_defs}
                    ]{
  cells={valign=m},
  colspec = {|X[l,.20\linewidth]| X[l]|},}
  \hline
   \SetCell[c=1]{c}{\textbf{Fallacy}} & \SetCell[c=1]{c}{\textbf{Formal Definition}} \\
    \hline
Ad Hominem & E claims P. E's character is attacked (A). Therefore, negate P. \\
\cline{1-2}
    Tu Quoque & E claims P, but E is acting as if E negates P. Therefore, negate P. \\
    \cline{1-2}
    Guilt by Association & E1 claims P. Also, E2 claims P, and E2's character is attacked (A). Therefore, negate P. OR, E1 claims P. E2's character is attacked (A) and is similar to E1. Therefore, negate P. \\
    \cline{1-2}
    Ad Populum & A lot of people believe/do P. Therefore, P. OR only a few people believe/do P. Therefore, negate P. \\
    \cline{1-2}
    Appeal to Nature & P1 is natural. P2 is not natural. Therefore, P1 is better than P2. OR P1 is natural, therefore P1 is good. \\
    \cline{1-2}
    Appeal to Tradition & We have been doing P for generations. Therefore, we should keep doing P. OR our ancestors thought P. Therefore, P. \\
    \cline{1-2}
    Appeal to False Authority & E claims P (when E is seen as an authority on the facts relevant to P). Therefore, P. \\
    \cline{1-2}
    Causal Oversimplification & P1 caused C (although P2, P3, P4, etc. also contributed to C). \\
    \cline{1-2}
    Hasty Generalization & Sample E1 is taken from population E. (Sample E1 is a very small part of population E.) Conclusion C is drawn from sample E1. \\
    \cline{1-2}
    False Causality & P is associated with C (when the link is mostly temporal and not logical). Therefore, P causes C. \\
    \cline{1-2}
    False Analogy & E1 is like E2. E2 has property P. Therefore, E1 has property P. (but E1 really is not too much like E2) \\
    \cline{1-2}
    False Dilemma & Either P1 or P2, while there are other possibilities. OR either P1, P2, or P3, while there are other possibilities. \\
    \cline{1-2}
    Slippery Slope & P1 implies P2, then P2 implies P3,... then C which is negative. Therefore, negate P1. \\
    \cline{1-2}
    Fallacy of Division & E1 is part of E, E has property P. Therefore, E1 has property P. \\
    \cline{1-2}
    Straw Man & E1 claims P. E2 restates E1's claim (in a distorted way P'). E2 attacks (A) P'. Therefore, negate P. \\
    \cline{1-2}
    Circular Reasoning & C because of P. P because of C. OR C because C. \\
    \cline{1-2}
    Equivocation & No logical form: P1 uses a term T that has a meaning M1. P2 uses the term T with the meaning M2 to mislead. \\
    \cline{1-2}
    Appeal to Positive Emotion & P is positive. Therefore, P. \\
    \cline{1-2}
    Appeal to Anger & E claims P. E is outraged. Therefore, P. Or E1 claims P. E2 is outraged by P. Therefore, P (or negate P depending on the situation). \\
    \cline{1-2}
    Appeal to Fear & If negate P1, something terrible P2 will happen. Therefore, P1. \\
    \cline{1-2}
    Appeal to Pity & P which is pitiful, therefore C, with only a superficial link between P and C. \\
    \cline{1-2}
    Appeal to Ridicule & E1 claims P. E2 makes P look ridiculous, by misrepresenting P (P'). Therefore, negate P. \\
    \cline{1-2}
    Appeal to Worse Problems & P1 is presented. P2 is presented as a best-case. Therefore, P1 is not that good. OR P1 is presented. P2 is presented as a worst-case. Therefore, P1 is very good. \\
    \hline
\end{longtblr}
}

\section{Prompt Templates}
\begin{table*}[htbp]
  \centering
  \resizebox{\linewidth}{!}{
    \begin{NiceTabular}{|p{2cm}|c|p{58em}|}
    \toprule
    \multicolumn{1}{l}{\textbf{Scheme}} & \multicolumn{1}{c}{\textbf{Round}} & \multicolumn{1}{c}{\textbf{Prompt Template}} \\
    \midrule
    \multicolumn{1}{l}{WD} & Single & Given \textcolor{red}{[\#Classes]} types of fallacies, namely, \textcolor{red}{[Fallacy List]}, and \textcolor{red}{<Discourse Type>} below, determine whether any / which of the fallacies given is present in \textcolor{red}{<Discourse Segment to be\newline{}Classified>}?\newline{}\textcolor{red}{[Discourse]}\newline{}Output your answer in JSON format \{"fallacy": name\_of\_the\_fallacy, "explanation":\newline{}in\_a\_sentence\_or\_two.\} (If none of the fallacies is found, output \{"fallacy": "No Fallacy",\newline{}"explanation": in\_a\_sentence\_or\_two\}). Only output JSON. \\
    \midrule
    \multicolumn{1}{l}{WD} & Single & Based on the following definitions of fallacies, \textcolor{red}{[Fallacy Definitions]}, given \textcolor{red}{<Discourse Type>} below, determine whether any / which of the fallacies defined above is present in \textcolor{red}{<Discourse}\newline{}\textcolor{red}{Segment to be Classified>}?\newline{}\textcolor{red}{[Discourse]}\newline{}Output your answer in JSON format \{"fallacy": name\_of\_the\_fallacy, "explanation":\newline{}in\_a\_sentence\_or\_two.\} (If none of the fallacies is found, output \{"fallacy": "No Fallacy",\newline{}"explanation": in\_a\_sentence\_or\_two\}). Only output JSON. \\
    \midrule
    \multicolumn{1}{l}{\multirow{2}[2]{*}{CoT}} & R1    & Given the following \textcolor{red}{<Discourse Type>},\newline{}\textcolor{red}{[Discourse]} \newline{}and the following \textcolor{red}{[\#Fallacy]} types of fallacies, namely, \textcolor{red}{[Fallacy List]}, which of the listed fallacies is present in \textcolor{red}{<Discourse Segment to be Classified>}? \textcolor{green}{Now, let’s think step by step.} \\
    \cmidrule{2-3}
          & R2    & Output your previous conclusion in JSON format \{"fallacy": name\_of\_the\_fallacy. Only output JSON.\} \\
    \midrule
    \multicolumn{1}{l}{\multirow{2}[2]{*}{DG}} & R1    & Give a definition to each of the following \textcolor{red}{[\#Fallacy]} types of fallacies in \textcolor{red}{[Fallacy List]}. \\
    \cmidrule{2-3}
          & R2    &  Based on the definitions you provided, given the \textcolor{red}{<Discourse Type>} below, determine which of the defined fallacies is present in \textcolor{red}{<Discourse Segment to be Classified>}? \newline{}\textcolor{red}{<Discourse Type>: [Discourse]}\newline{}Output your answer in JSON format \{"fallacy": name\_of\_the\_fallacy. Only output JSON.\} \\
    \midrule
    \multicolumn{1}{l}{\multirow{2}[2]{*}{GFA}} & R1    & Given the following \textcolor{red}{<Discourse Type>}, is the \textcolor{red}{<Discourse Segment to be Classified>} logically reasonable or potentially fallacious? \newline{}\textcolor{red}{<Discourse Type>}: \textcolor{red}{[Discourse]}\newline{}Give your analysis. \\
    \cmidrule{2-3}
          & R2    & According to your previous analysis, considering \textcolor{red}{[\#Fallacy]} of fallacies \textcolor{red}{[Fallacy List]}, determine which of these listed fallacies is present in the \textcolor{red}{<Discourse Segment to be Classified>}?\newline{}Output your answer in JSON format \{"fallacy": name\_of\_the\_fallacy. Only output JSON.\} \\
    \midrule
    \multicolumn{1}{l}{\multirow{3}[2]{*}{\newline{}GFA-W}} & R1    & \multicolumn{1}{l}{(\textsc{Logic}) Extract and summarize the focal argument in the following segment of discourse: \textcolor{red}{[Discourse]}} \\
    \cmidrule{2-3}
          & R2    & Is \textcolor{red}{<Discourse Segment to be Classified>} logically reasonable or potentially fallacious? Give your\newline{}analysis. \\
    \cmidrule{2-3}
          & R3    & According to your previous analysis, considering \textcolor{red}{[\#Fallacy]} of fallacies \textcolor{red}{[Fallacy List]}, determine which of these listed fallacies is present in the \textcolor{red}{<Discourse Segment to be Classified>}?\newline{}Output your answer in JSON format \{"fallacy": name\_of\_the\_fallacy. Only output JSON.\} \\
    \midrule
    \multicolumn{1}{l}{\multirow{3}[2]{*}{P\&C$^{1}$}} & R1    & Given the following \textcolor{red}{<Discourse Type>}\newline{}\textcolor{red}{[Discourse]}\newline{}extract and summarize \textcolor{red}{<Discourse Argument>} by pointing out the premise(s) and conclusion of the argument. \\
    \cmidrule{2-3}
          & R2    & A fallacy is defined as an argument where the premises do not entail the conclusion. According to this definition, is \textcolor{red}{<Discourse Segment to be Classified>} logically reasonable or potentially fallacious? Give your analysis. \\
    \cmidrule{2-3}
          & R3    & According to your previous analysis, considering \textcolor{red}{[\#Fallacy]} of fallacies \textcolor{red}{[Fallacy List]}, determine which of these listed fallacies is present in the \textcolor{red}{<Discourse Segment to be Classified>}?\newline{}Output your answer in JSON format \{"fallacy": name\_of\_the\_fallacy. Only output JSON.\} \\
    \midrule
    \multicolumn{1}{l}{\multirow{3}[2]{*}{P\&C$^{2}$}} & R1    & Given the following \textcolor{red}{<Discourse Type>}\newline{}\textcolor{red}{[Discourse]}\newline{}extract and summarize \textcolor{red}{<Discourse Argument>} by pointing out the premise(s) and conclusion of the argument. \\
    \cmidrule{2-3}
          & R2    & Whether or not the premise(s) of \textcolor{red}{<Discourse Argument>} entail the conclusion? Give\newline{}your analysis. \\
    \cmidrule{2-3}
          & R3    & A fallacy is defined as an argument where the premises do not entail the conclusion. According to your previous analysis, considering \textcolor{red}{[\#Fallacy]} of fallacies \textcolor{red}{[Fallacy List]}, determine which of these listed fallacies is present in the \textcolor{red}{<Discourse Segment to be Classified>}?\newline{}Output your answer in JSON format \{"fallacy": name\_of\_the\_fallacy. Only output JSON.\} \\
    \bottomrule
    \end{NiceTabular}%
  }
  \vspace{-5pt}
  \caption{Sample templates of our proposed single-round and multi-round prompting schemes. Notations: \textbf{DG} for Definition Generation. \textbf{GFA} for General Fallacy Analysis. \textbf{GFA-W} for General Fallacy Analysis with Warm up. \textbf{P\&C} for Premises \& Conclusion. \textbf{CoT} for Zero-shot CoT.
  }
  \label{tab:prompt_template}%
\end{table*}%

\newpage
\section{Error Analysis}
\subsection{Performance Analysis of Premises \& Conclusion}
\label{sec:perform_analysis_pec}
We compare the confusion matrix of GPT-4 under the top three best performed prompting schemes and the confusion matrix of GPT-4 under the multi-round prompting scheme of Premises \& Conclusion on each dataset. We find that, under the Premises \& Conclusion scheme, GPT-4 is significantly biased to over predict ``No Fallacy'' when this label is given or fallacy types that contain words related to the semantics of ``causality'', such as``causal'' and ``generalization''. We speculate that the prompt formulation of Premises \& Conclusion by offering the formal definition of the term fallacy ``{\it whether the premises entail the conclusion}'' has a misleading implication that renders GPT-4 to place excessive attention on fallacy conceptions with semantics that overlap with the word ``entail''. This is why Premises \& Conclusion prompting scheme underperforms the other prompting methods.
\begin{figure}[htbp]
  \centering
  \includegraphics[width=1\linewidth]{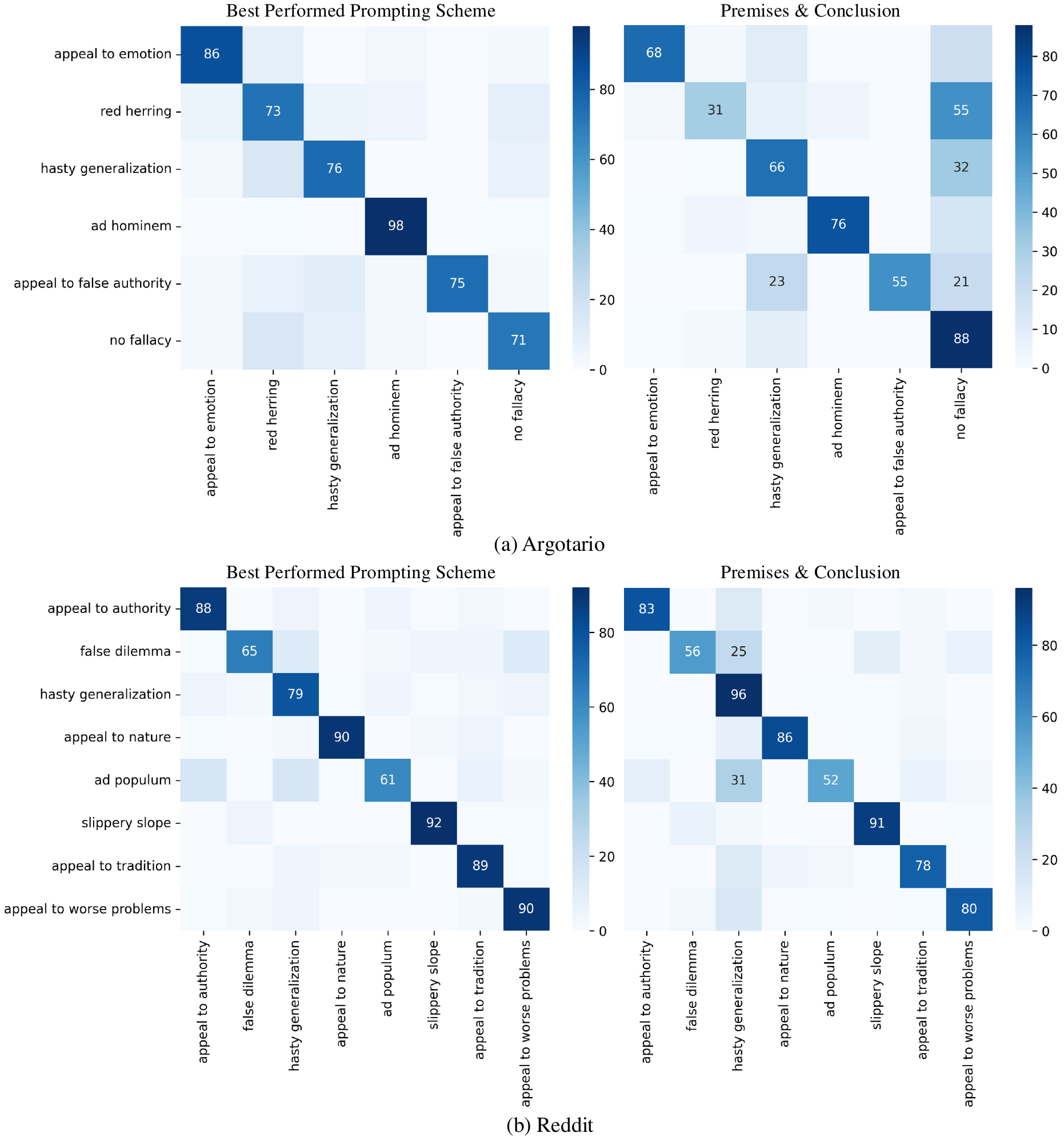}
  \caption{{Comparison of GPT-4's results under the best performed prompting scheme and the multi-round Premise \& Conclusion on \textsc{Argotario} and \textsc{Reddit}.}
  }
  \label{fig:error_analysis_argotario_reddit}
  \vspace{-10pt}
\end{figure}

\begin{figure}[htbp]
  \centering
  \includegraphics[width=1\linewidth]{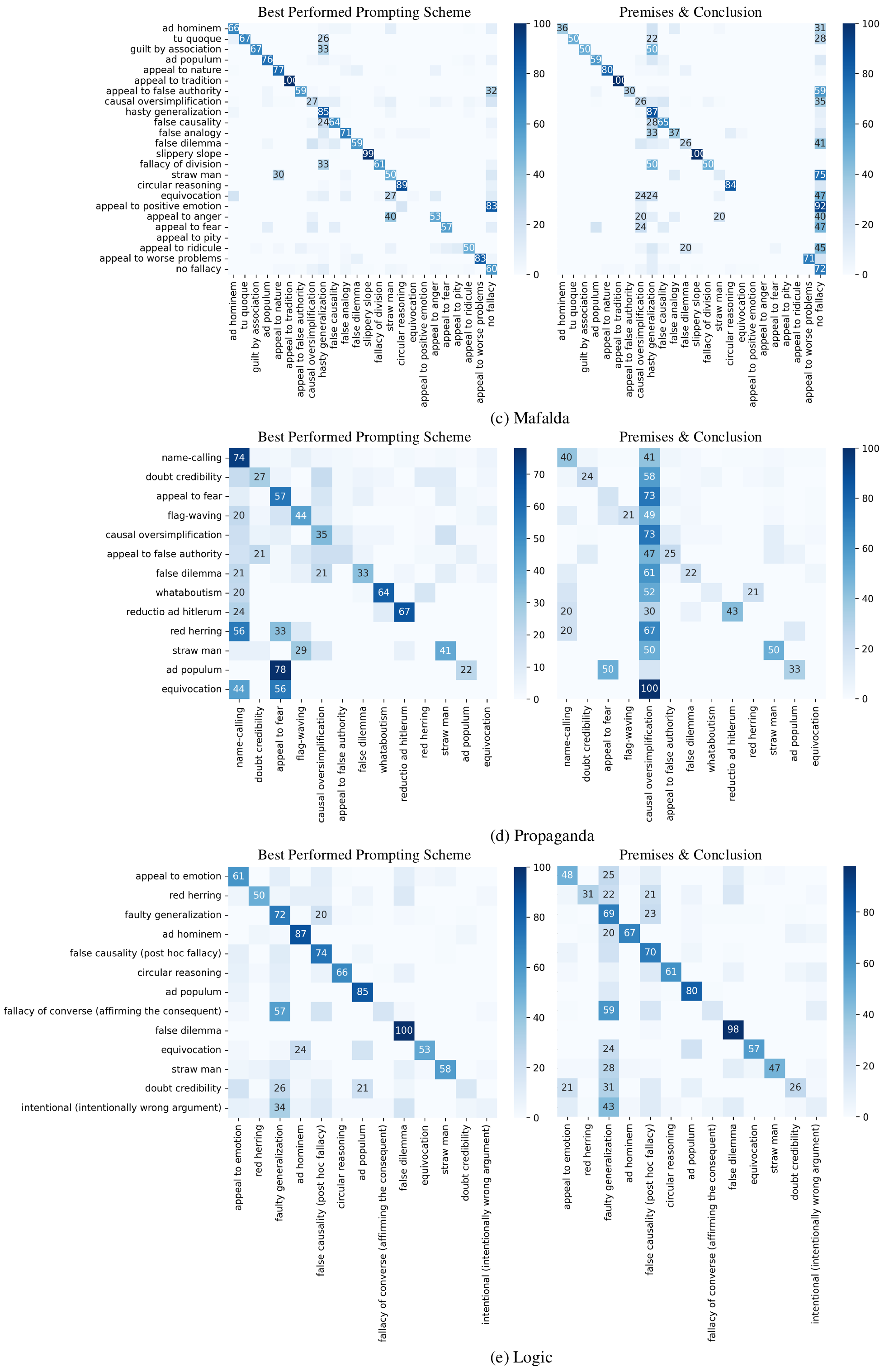}
  \caption{{Comparison of GPT-4's results under the best performed prompting scheme and the multi-round Premise \& Conclusion on \textsc{Mafalda}, \textsc{Propaganda} and \textsc{Logic}.}
  }
  \label{fig:error_analysis_mafalda_propaganda_logic}
  \vspace{-10pt}
\end{figure}

\begin{figure}[htbp]
  \centering
  \includegraphics[width=1\linewidth]{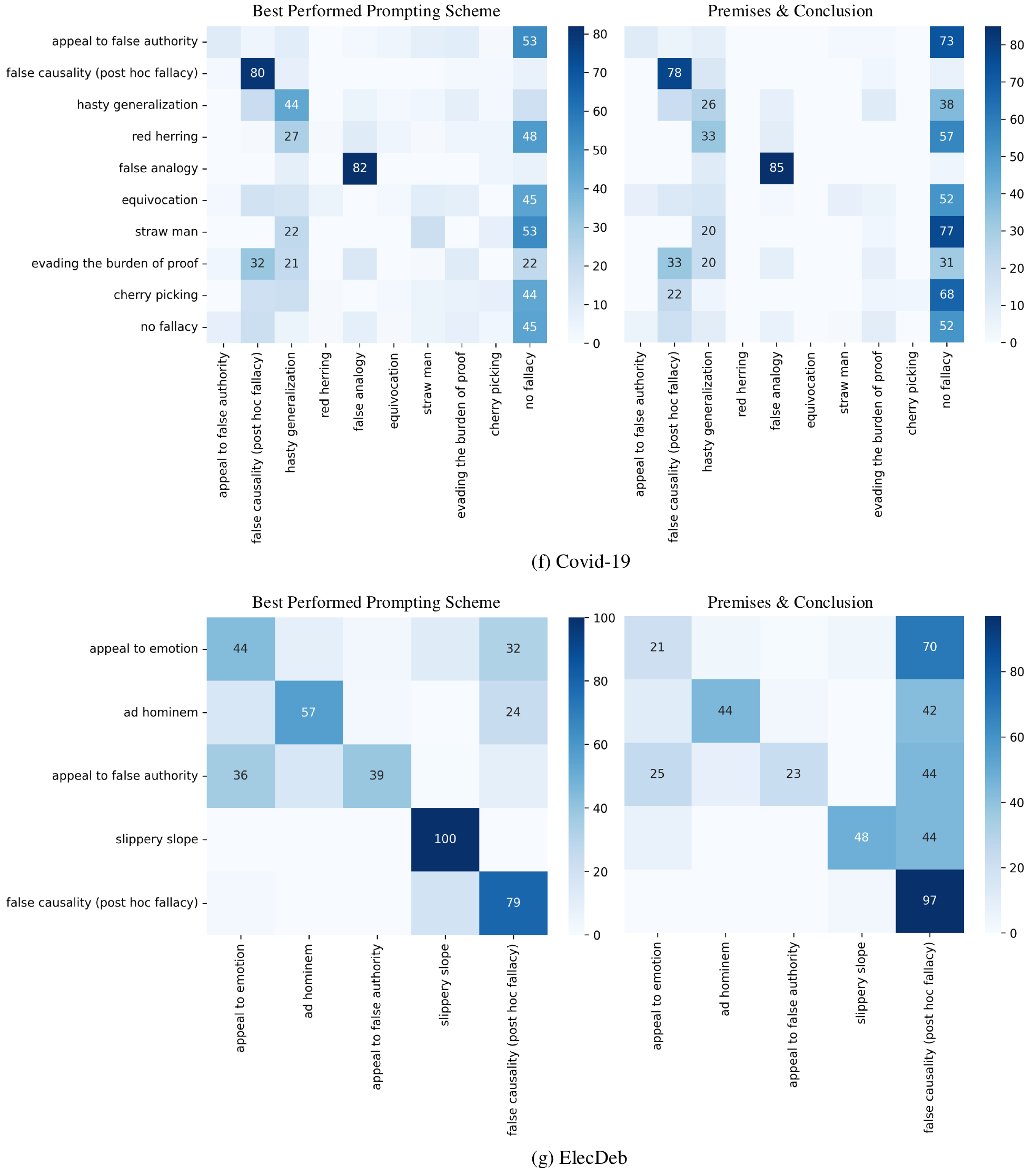}
  \caption{{Comparison of GPT-4's results under the best performed prompting scheme and the multi-round Premise \& Conclusion on \textsc{Covid-19} and \textsc{ElecDeb}.}
  }
  \label{fig:error_analysis_covid_elecdeb}
  \vspace{-10pt}
\end{figure}

\clearpage

\subsection{Error Cases}
  {
  \footnotesize
  \tabcolsep=2pt
  \centering
    \begin{longtable}{|c|c|c|p{40em}|}
        \hline
    \multicolumn{1}{|c|}{\textbf{Scheme}} & \multicolumn{1}{c|}{\textbf{Model}} &       & \multicolumn{1}{r|}{} \\
    \hline
    \multirow{8}[16]{*}{DG} & \multirow{4}[8]{*}{GPT-4} & \textbf{Input} & A: Do we need a global environment task force? B: The global taskforce for human rights is doing a good job. That's why we need one for the environment, too. \\
\cline{3-4}          &       & \textbf{Output} & Hasty Generalization: A conclusion based on insufficient or biased evidence; rushing to a conclusion before you have all relevant facts. \\
\cline{3-4}          &       & \textbf{Pred.} & No Fallacy \\
\cline{3-4}          &       & \textbf{GT} & Hasty Generalization \\
\cline{2-4}          & \multirow{4}[8]{*}{Mistral} & \textbf{Input} & <YouTube conspiracy theorist> Jake Morphonios said that for all he knows, Doug Haig didn't play a role in Paddock's massacre.  \\
\cline{3-4}          &       & \textbf{Output} & Red Herring: Introducing a distracting or irrelevant topic, so as to lead others away from the main subject at hand, and avoid addressing the actual argument.\newline{}Name-calling Using derogatory or offensive terms to describe someone or their arguments, instead of presenting a counterargument. \\
\cline{3-4}          &       & \textbf{Pred.} & Red Herring \\
\cline{3-4}          &       & \textbf{GT} & Name-calling \\
    \hline
    \multirow{8}[16]{*}{GFA} & \multirow{4}[8]{*}{GPT-3.5} & \textbf{Input} & something that kills is deadly. \\
\cline{3-4}          &       & \textbf{Output} & The focal argument of the discourse is logically reasonable. The statement ``something that kills is deadly'' is a tautology, as it simply states that if something has the capability to kill, it is indeed deadly. This argument is logically sound and does not contain any fallacies. \\
\cline{3-4}          &       & \textbf{Pred.} & No Fallacy \\
\cline{3-4}          &       & \textbf{GT} & Circular Reasoning \\
\cline{2-4}          & \multirow{4}[8]{*}{Llama-2} & \textbf{Input} & If you are outraged by this, stop eating meat and dairy. Otherwise, you are a hypocrite \\
\cline{3-4}          &       & \textbf{Output} & The comment's argument is potentially fallacious. The argument is an example of the ``hasty generalization'' fallacy, where a conclusion is drawn based on insufficient or incomplete evidence. The comment assumes that anyone who is outraged by the topic must be a hypocrite if they consume meat and dairy, without considering the possibility that they may have valid reasons for their outrage that have nothing to do with their dietary choices. Additionally, the comment does not address the possibility that some people may be vegetarian or vegan and still be outraged by the topic. Therefore, the comment's argument is not logically reasonable and may be considered fallacious. \\
\cline{3-4}          &       & \textbf{Pred.} & Hasty Generalization \\
\cline{3-4}          &       & \textbf{GT} & False Dilemma \\
    \hline
    \multirow{8}[16]{*}{GFA-W} & \multirow{4}[8]{*}{GPT-3.5} & \textbf{Input} & ``<Are the 2,000 U.S. troops in Syria really up to that task,> should the Turks decide to cleanse the Syrian border of Kurds, or should the Syrian regime decide to take back territory occupied by the Kurds?'' \\
\cline{3-4}          &       & \textbf{Output} & The fragment highlighted by `<>' is potentially fallacious, specifically exhibiting a form of hasty generalization. It questions whether 2,000 U.S. troops are sufficient to handle major conflicts without providing detailed evidence or context about the capabilities of these troops, their operational support, or the nature of the potential military engagements they might face. This can lead to an oversimplified conclusion about the effectiveness of U.S. military presence based on troop numbers alone, without considering other strategic factors. \\
\cline{3-4}          &       & \textbf{Pred.} & Causal Oversimplification \\
\cline{3-4}          &       & \textbf{GT} & Doubt Credibility \\
\cline{2-4}          & \multirow{4}[8]{*}{Mistral} & \textbf{Input} & Paige wants to play a guitar solo on our next recording. Her father is a horrible musician, though, so I doubt that Paula's any good either. \\
\cline{3-4}          &       & \textbf{Output} & The argument presented is potentially fallacious, known as ``Negative Ad Hominem Fallacy'', which is an argument that attacks the person rather than the argument, and incorrectly assumes that because the father is a ``horrible musician'', his daughter (Paige) must be equally "horrible" as well. It is not fair to judge the skills of a person (Paige) based on the skills of another person (her father) without considering Paige's own talents or skills. \\
\cline{3-4}          &       & \textbf{Pred.} & Red Herring \\
\cline{3-4}          &       & \textbf{GT} & Hasty Generalization \\
    \hline
    \multirow{8}[16]{*}{CoT} & \multirow{4}[8]{*}{GPT-4} & \textbf{Input} & ``I have contacted the school's principal, Dr.'',\newline{}``Branch, and I trust that he will take this matter seriously and deal with it appropriately.''\newline{}``Is the principal acting on this?'' \\
\cline{3-4}          &       & \textbf{Output} & The fallacy present in the segment highlighted by `<>' is ``Red Herring.'' This is because the question about the principal's action diverts attention from the main issue discussed in the article (Islamizing the schools) to the actions of the school principal, which is a separate issue. \\
\cline{3-4}          &       & \textbf{Pred.} & Red Herring \\
\cline{3-4}          &       & \textbf{GT} & Doubt Credibility \\
\cline{2-4}          & \multirow{4}[8]{*}{Llama-3} & \textbf{Input} & I don't think you can just rely on U.N. sanctions, Randee. But you're absolutely correct, it is a threat, it's a huge threat. \\
\cline{3-4}          &       & \textbf{Output} & After analyzing the speech, I found that the fallacy present in the highlighted argument is the False Causality (Post Hoc Fallacy). The speaker is implying that the threat from North Korea grew while the President was preoccupied with Iraq, suggesting that the President's focus on Iraq caused the threat to grow. However, this is a post hoc fallacy because it's an assumption that the President's focus on Iraq caused the threat to grow, without providing any evidence or logical connection between the two. \\
\cline{3-4}          &       & \textbf{Pred.} & False Causality \\
\cline{3-4}          &       & \textbf{GT} & Appeal to Emotion \\
    \hline
    \multirow{8}[16]{*}{P\&C} & \multirow{4}[8]{*}{GPT-4} & \textbf{Input} & I bought a ticket to win a new car at the mall, since I have never won anything like that in the past."", \\
\cline{3-4}          &       & \textbf{Output} & The premise that the person has never won anything like a new car in the past does not logically entail the conclusion that they bought a ticket to win a new car at the mall. The premise provides a context or a possible motivation for why they might have decided to buy a ticket, perhaps hoping to change their luck, but it does not necessarily lead to the conclusion that buying a ticket was a rational or necessary action. There is no causal or logical necessity connecting the lack of previous winnings to the decision to buy a ticket; the conclusion does not directly follow from the premise. \\
\cline{3-4}          &       & \textbf{Pred.} & No Fallacy \\
\cline{3-4}          &       & \textbf{GT} & False Causality \\
\cline{2-4}          & \multirow{4}[8]{*}{Llama-3} & \textbf{Input} & A: Is Turkey ready to join the European Union? \newline{}B: Turkey helped us with the Refugees. They were beneficial for the EU there, they will be beneficial for the EU in general. \\
\cline{3-4}          &       & \textbf{Output} & The premise ``Putin has nothing to say regarding the EU. It is our choice who joins and who does not'' does not entail the conclusion ``Turkey is ready to join the European Union''. This premise only establishes that the decision to admit Turkey into the EU is up to the EU member states, not Putin, and does not provide any information about Turkey's readiness to join. The premise ``Turkey and Europe share much culture and many people from Turkey already live in European countries. They are ready'' is more relevant to the conclusion, but it is still a weak argument. The fact that Turkey and Europe share culture and have people living in each other's countries does not necessarily mean that Turkey is ready to join the EU. There are many other factors to consider, such as Turkey's political system, economic stability, and human rights record, among others. Therefore, the premises do not entail the conclusion, and B's argument is flawed. \\
\cline{3-4}          &       & \textbf{Pred.} & Appeal to Consequences \\
\cline{3-4}          &       & \textbf{GT} & Hasty Generalization \\
    \hline
    \caption{Error cases of GPT-4 and Llama3-Chat (8B) for each multi-round prompting scheme.}
    \end{longtable}%
  }

\subsection{Misclassification Confusion Matrices of GPT-4 and Llama3}
\label{sec:confusion_matrix_details}
\begin{figure}[htbp]
  \centering
  \includegraphics[width=\linewidth]{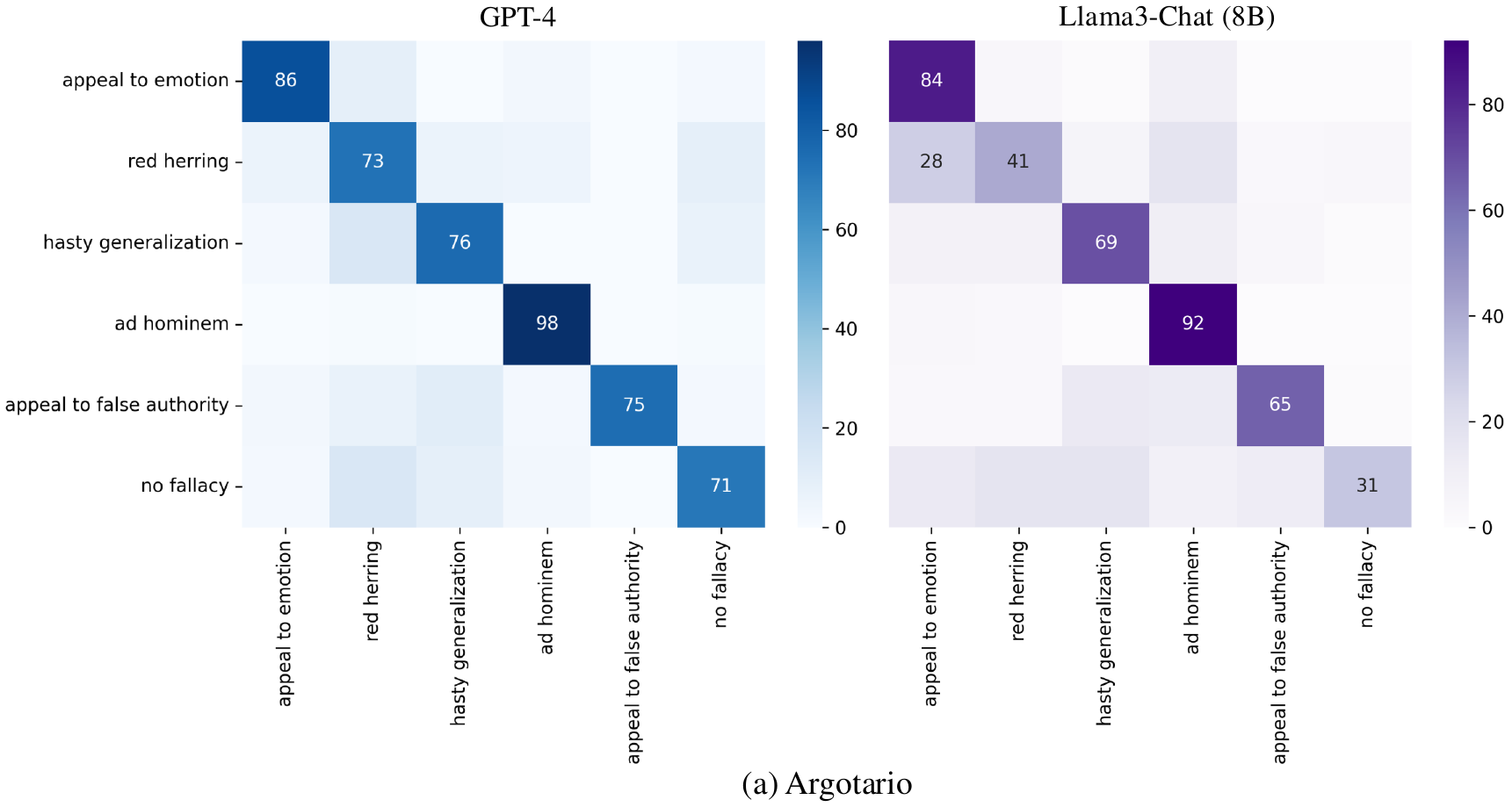}
  \caption{Misclassification confusion matrix of GPT-4 and Llama3-Chat (8B) on {\sc Argotario}. Cell values are the percentages of row fallacy types that are misclassified as column fallacy types.
  }
  \label{fig:cm_argotario}
  \vspace{-10pt}
\end{figure}

\begin{figure}[htbp]
  \centering
  \includegraphics[width=0.95\linewidth]{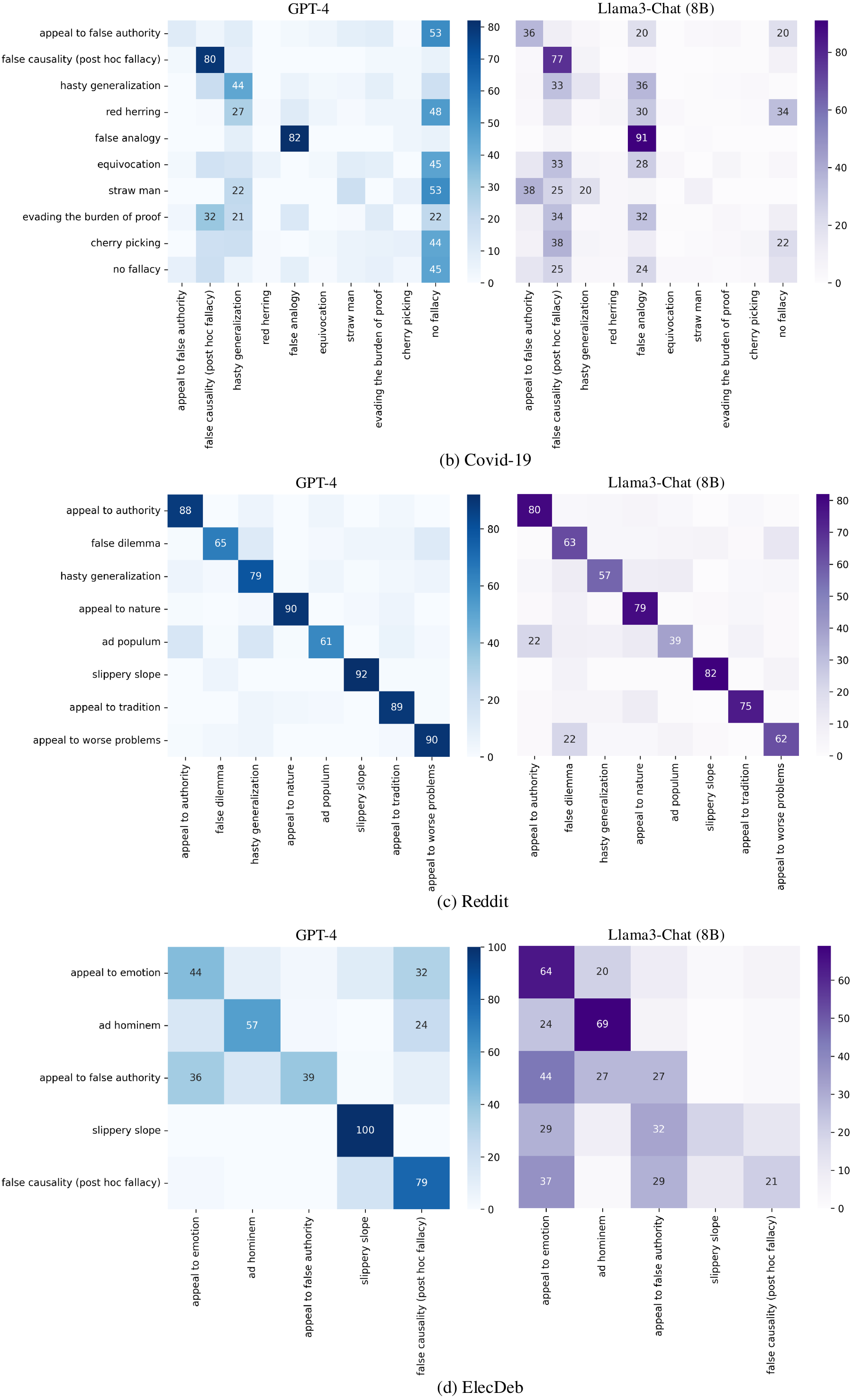}
  \caption{Misclassification confusion matrix of GPT-4 and Llama3-Chat (8B) on {\sc Covid-19}, \textsc{Reddit} and \textsc{ElecDeb}. Cell values are the percentages of row fallacy types that are misclassified as column fallacy types.
  }
  \label{fig:cm_covid_reddit_elecdeb}
  \vspace{-10pt}
\end{figure}

\begin{figure}[htbp]
  \centering
  \includegraphics[width=\linewidth]{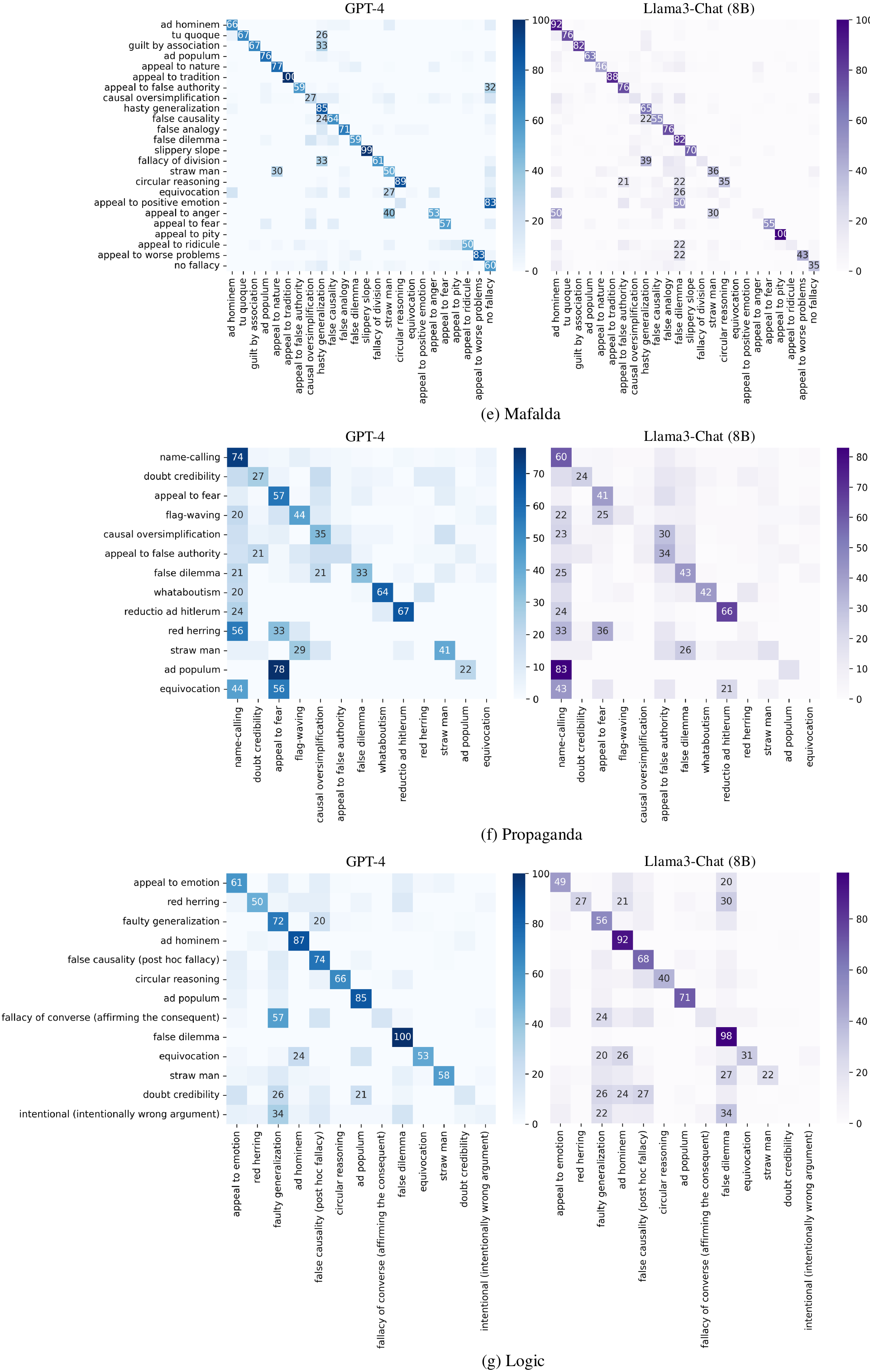}
  \caption{Misclassification confusion matrix of GPT-4 and Llama3-Chat (8B) on {\sc Mafalda}, \textsc{Propaganda} and \textsc{Logic}. Cell values are the percentages of row fallacy types that are misclassified as column fallacy types.
  }
  \label{fig:cm_mafalda_propaganda_logic}
  \vspace{-10pt}
\end{figure}

\end{document}